\documentclass{article}

\usepackage{microtype}
\usepackage{graphicx}
\usepackage{subfigure}
\usepackage{booktabs} %

\usepackage{hyperref}

\usepackage{amsmath}
\usepackage{amssymb}
\usepackage{mathtools}
\usepackage{amsthm}

\usepackage[capitalize,noabbrev]{cleveref}

\usepackage{url}            %
\usepackage{amsfonts}       %
\usepackage{nicefrac}       %
\usepackage{amsmath}
\usepackage{color}
\usepackage{standalone}
\usepackage{tikz}
\usepackage{setspace}
\usepackage{bm}
\usepackage{fullpage}
\usepackage{stmaryrd}
\usepackage{dsfont}
\usepackage{enumitem}
\usepackage{bbm}
\usepackage{float}

\usepackage{fancyhdr}
\usepackage{xcolor} %
\usepackage{algorithm}
\usepackage{algorithmic}
\usepackage{natbib}
\usepackage{eso-pic} %
\usepackage{forloop}
\usepackage{url}

\newcommand{\TODO}[1]{%
\ifmmode
\text{\textcolor{red}{TODO: #1}}
\else
\textcolor{red}{TODO: #1}
\fi
}

\usetikzlibrary{decorations.pathreplacing,calc}

\usepackage{amsthm}
\usepackage{thmtools}
\usepackage{thm-restate}

\renewcommand{\tilde}{\widetilde}
\renewcommand{\hat}{\widehat}

\DeclareMathOperator*{\argmin}{arg\,min}

\newcommand{\PP}{\mathbb{P}}
\renewcommand{\Pr}{\PP}

\newcommand{\E}{\mathbb{E}}

\newcommand{\yhatt}{\hat{y}_{t}}
\newcommand{\Epsi}{\mathcal{E}}
\newcommand{\bphat}{\hat{\p}}

\newcommand{\sumK}{\sum_{j=1}^K}
\newcommand{\sumT}{\sum_{t=1}^T}
\newcommand{\sumTN}{\sum_{t=N+1}^T}

\newcommand{\id}{\mathbbm{1}}
\newcommand{\sign}{\textnormal{sign}}

\newcommand{\Fset}{\mathcal{F}}
\newcommand{\Cset}{\mathcal{C}}

\newcommand{\half}{\tfrac{1}{2}}

\newcommand{\Cshat}{{\Cset}}
\newcommand{\Phat}{{\mathcal{P}}}
\newcommand{\phat}{\hat{p}}

\newcommand{\OPT}{\mbox{\textsc{opt}}}

\newcommand{\scO}{\mathcal{O}}

\renewcommand{\log}{\ln}

\DeclareBoldMathCommand{\p}{p}
\DeclareBoldMathCommand{\z}{z}
\DeclareBoldMathCommand{\v}{v}
\DeclareBoldMathCommand{\w}{w}
\DeclareBoldMathCommand{\W}{W}
\DeclareBoldMathCommand{\y}{y}
\DeclareBoldMathCommand{\h}{h}
\DeclareBoldMathCommand{\q}{q}
\DeclareBoldMathCommand{\e}{e}
\DeclareBoldMathCommand{\b}{b}
\DeclareBoldMathCommand{\g}{g}
\DeclareBoldMathCommand{\u}{u}
\DeclareBoldMathCommand{\U}{U}
\DeclareBoldMathCommand{\L}{L}
\DeclareBoldMathCommand{\l}{l}
\DeclareBoldMathCommand{\1}{1}
\DeclareBoldMathCommand{\0}{0}
\DeclareBoldMathCommand{\a}{a}
\DeclareBoldMathCommand{\x}{x}
\DeclareBoldMathCommand{\c}{c}
\DeclareBoldMathCommand{\I}{I}

\theoremstyle{plain}
\newtheorem{theorem}{Theorem}[section]

\newtheorem{lemma}[theorem]{Lemma}

\theoremstyle{definition}

\theoremstyle{remark}

\usepackage[textsize=tiny]{todonotes}

\newcommand*\samethanks[1][\value{footnote}]{\footnotemark[#1]}

\title{Trading-Off Payments and Accuracy \\ in Online Classification with Paid Stochastic Experts}
\author{Dirk van der Hoeven\thanks{Equal Contribution. Correspondence to: Dirk van der Hoeven
\url{dirk@dirkvanderhoeven.com}}$^{\,\,\,}$\thanks{Korteweg-de Vries Institute for Mathematics, University of Amsterdam, Amsterdam, The Netherlands}
\and Ciara Pike-Burke\samethanks[1] $^{\,\,}$\thanks{Department of Mathematics, Imperial College London, London, UK}
\and Hao Qiu\thanks{Università degli Studi di Milano, Milan, Italy} 
\and Nicolò Cesa-Bianchi\samethanks[4]$^{\,\,}$ \thanks{ Politecnico di Milano, Milan, Italy} }

\begin{document}
\maketitle

\begin{abstract}
We investigate online classification with paid stochastic experts. Here, before making their prediction, each expert must be paid. The amount that we pay each expert directly influences the accuracy of their prediction through some unknown Lipschitz ``productivity'' function. In each round, the learner must decide how much to pay each expert and then make a prediction. They incur a cost equal to a weighted sum of the prediction error and upfront payments for all experts. We introduce an online learning algorithm whose total cost after $T$ rounds exceeds that of a predictor which knows the productivity of all experts in advance by at most $\mathcal{O}\big(K^2(\ln T)\sqrt{T}\big)$ where $K$ is the number of experts.
In order to achieve this result, we combine Lipschitz bandits and online classification with surrogate losses. These tools allow us to improve upon the bound of order $T^{2/3}$ one would obtain in the standard Lipschitz bandit setting. Our algorithm is empirically evaluated on synthetic data.
\end{abstract}

\section{Introduction}\label{sec:introduction}
We investigate online classification in the framework of prediction with expert advice where, in each round, the learning agent predicts an unknown binary label by aggregating the stochastic predictions of a number of experts. At the end of each round, the learner observes the true label and updates the function used to aggregate experts. In the variant considered in this work, we assume that at the beginning of a round the learner allocates a payment to each expert
which affects the expert's performance in that round.
This payment model of expert advice is realistic in many scenarios since human annotators will often only give useful advice if they are adequately compensated, and machine annotators may require more computation to return accurate predictions. Moreover, monetary incentives have been studied in crowdsourcing \citep{ho2015incentivizing,ho2016adaptive}. Although this is a different setting to that considered here, it is natural to study the effect of these payments in online binary classification with stochastic expert advice.

Motivated by results in crowdsourcing---e.g., \cite{ho2016adaptive}---we assume that each expert has a \emph{productivity function} which determines the probability that they predict the label correctly given the payment they received. The productivity function can be different for each expert and is initially unknown to the learner.
In each round, the learner pays each expert $j=1,\dots, K$ some amount $c_j \in [0,1]$ before observing their advice. The accuracy of the advice that expert $j$ returns depends on the amount they are paid through the unknown productivity function, $p_j : [0,1] \to [0,1]$, where $p_j(c)$ is the probability that expert $j$ is correct when the payment is $c$. %
The learner can use the expert advice to improve their prediction, but they also want to minimize the payments to the experts. Therefore, they must trade-off between the price of the expert advice, and any improvements to prediction accuracy it may bring.

We define the learner's cost over a sequence of $T$ rounds as the sum of classification mistakes and payments to experts.
If the probabilities $p_j(c_j)$ are known for some $c_1,\ldots,c_K$ and for all experts $j=1,\dots, K$, then we can write down the Bayes-optimal cost. In particular, if the events that each expert makes a mistake are independent,
then the error probability of the Bayes-optimal aggregated prediction is known to decrease exponentially fast with exponent $-2\big(\gamma_1^2(c_1)+\cdots+\gamma_K^2(c_K)\big)$, where $\gamma_j(c_j) = \big(p_j(c_j)-\half\big)$ for $j=1,\ldots,K$ \citep{ho2013adaptive}. Given productivity functions $\p = \big(p_1,\ldots,p_K\big)$, we then define the
optimal cost over $T$ rounds by $T\OPT(\p)$, where
\[
    \OPT(\p) = \min_{c_1,\dots, c_K \in [0,1]}\left\{ e^{-2\sum_{j=1}^K \gamma_j(c_j)^2} + \lambda \sum_{j=1}^K c_j\right\}
\]
and $\lambda >0$ is a given parameter introduced to balance the trade-off between accuracy and payments. 

In this paper, we consider the case when $\p$ is unknown and the learner's goal is to minimize the regret,
\begin{align}
    R_T = \sum_{t=1}^T \left( \Pr_Z\big(\yhatt \neq y_t\big) + \lambda \sum_{j=1}^K c_{t,j} \!\right) - T\OPT(\p) \label{eqn:regdef}
\end{align}
where, in each round $t$, $\yhatt\in\{-1,1\}$ is the learner's prediction (which depends on the stochastic expert advice $Z$), $y_t \in \{-1,1\}$ is the adversarially chosen true label, and $c_{t,j}$ is the payment for expert $j$. Following the standard online learning protocol, we assume the true label $y_t$ is revealed at the end of each round. The learner can then use $y_t$ and the expert advice to learn the productivity functions.

Regret minimization in this problem presents several challenges. The first is the need to trade-off cost and accuracy while simultaneously learning how the payments affect the accuracy. Indeed, we only observe the predictions of the experts with the payments that we pay them. This introduces an additional exploration-vs-exploitation trade-off as is typically seen in bandit problems. However, as we discuss in Section~\ref{sec:prelim}, this is more challenging than in standard bandit problems since the relationship between the payments that we choose and the regret is more complex and possibly non-smooth. %

A further significant challenge is to combine the predictions from all experts when we only have estimates of their accuracy. 
In particular, if we have estimated an experts productivity function $\hat p_{t,j}(c_j)$, as being close to 0 or 1 for a specific $c_j$ in round $t$, directly using a majority or weighted majority aggregation approach could lead to undesirable scaling of the regret with $1/\min\{\hat p_{t,j}(c_j),1-\hat p_{t,j}(c_j)\}$ which can be arbitrarily large. 

Our approach to overcome these issues is based on a combination of several ideas. First, we discretize the payment interval and for each payment and expert combination we estimate the probability of a correct classification. To deal with the exploration vs exploitation challenge we rely on the optimism in the face of uncertainty principle---see, for example, \citep[Chapter~7]{lattimore2020bandit}. While this is a standard approach for stochastic losses, if it were to be directly applied to the discretized payments here, it would lead to an undesirable regret bound of $\scO(T^{2/3})$.\footnote{
To see why standard methods give a $T^{2/3}$ rate consider a simplified setting with one expert. Then, the mistake probability is equivalent to the productivity function of that expert, denoted by $p$. Since we assume that $p$ is Lipschitz, the problem is now reduced to Lipschitz bandits, for which a $T^{2/3}$ bound is known to be unavoidable \citep{slivkins2019introduction}. 
}
Instead, we combine the optimistic principle with tools from online classification with surrogate losses to obtain a $\scO(\sqrt{T})$ regret bound. Specifically, we use the randomized predictions of \citet{VanderHoeven2020exploiting, vanderHoeven2021beyond} which gains us considerable leeway in the analysis,
see Section~\ref{sec:analysis} for the details.

To avoid regret that scales with $1/\min\{p_j(c),1-p_j(c)\}$, we propose a modified aggregation approach. This approach simply follows the advice of one expert if we believe they are very likely to be correct (or wrong, in which case we use the opposite of their prediction). This aggregation approach allows us to have \emph{multiplicative} control over estimation errors rather than additive control. Combined with the
randomized prediction technique, multiplicative control over the estimation error is a crucial element in our analysis, and one of the major differences compared to standard analysis of aggregated classifiers. 

Combining these ideas with tight error bounds on the productivity function allows us to obtain the following result (implied by Theorem~\ref{th:mainO}). 
\begin{theorem}
The regret of LCB-GAPTRON (Algorithm~\ref{alg:nosigmoid}) satisfies
$ %
    R_T = \mathcal{O}\big(K^2(\ln T)\sqrt{T}\big)
$ %
where $K$ is the number of experts and $T$ is the number of rounds.
\end{theorem}

This result represents an improvement on the $T^{2/3}$ regret bound that would be achievable if we were to simply use an optimistic algorithm with discretized costs. We also demonstrate that our algorithm significantly outperforms this naive algorithm in several simulated environments. Our experiments also indicate that the most computationally demanding step of our algorithm can be replaced by a cheaper approximation with little impact on the regret.

\subsection{Related Work}
Online aggregation of experts is also studied in online boosting \citep{chen2012online,beygelzimer2015optimal}, a setting where there are no payments and the predictions of experts are adversarial.
When the average accuracy of the experts is unknown, \citet{beygelzimer2015optimal} prove a mistake bound of $\frac{8}{\gamma^2 K}T + \widetilde{\mathcal{O}}\big(\frac{K}{\gamma^2}\big)$, where $\frac{1}{2}-\gamma$ is an upper bound on the fraction of mistakes made by any expert. Note that, due to the adversarial assumption on the experts, the leading term in this bound vanishes at rate $K^{-1}$, as opposed to the exponential rate $e^{-K}$ achievable in our stochastic setting.

Our setting is also related to the framework of online prediction with limited expert advice \citep{seldin2014prediction,kale2014multiarmed}, where the predictions of experts and the payments are both chosen by an adversary. In this model, the learner can buy advice from any subset of experts at the price posted by the adversary. As the payments are not chosen by the learner, the trade-off between payments and accuracy is different from the one studied in this work.

Although our setting is online learning, solving classification tasks by aggregating the predictions of stochastic experts naturally arises also in the context of crowdsourced labeling \citep{zhang2016learning,vaughan2017making}. \citet{karger2014budget} study a setting where a requester has a set of $n$ homogeneous binary labeling tasks to be assigned to $m$ workers arriving in random order. Each worker $j$ is characterized by an unknown probability $p_j$ of labeling correctly any task and by a capacity $T_j$. At each round, the requester observes the capacity of the current worker and selects a subset of tasks respecting the worker's capacity. %
After the $m$ workers have been all assigned tasks, the requester aggregates their predictions to infer a single prediction for each task.

This model has been extended to consider heterogeneous tasks (so that each worker $j$ has a different accuracy $p_{i,j}$ for each task $i$) and costly access to ground truth labels for each task type \citep{ho2013adaptive}. \citet{tran2013efficient} also
extend the crowdsourcing model to a setting where %
workers have fixed and known costs and the requester must allocate tasks while respecting a budget constraint.

From an online learning viewpoint, these crowdsourcing papers consider a dual \textsl{pool-based} model, where a set of unlabeled points is preliminary given and experts arrive online. 
In contrast, in our problem the set of experts is fixed and new instances are considered in each round, thus the problem studied here is quite distinct from crowdsourcing.
In our work, we also consider payments that influence the workers' accuracy, which is not included in the classical crowdsourcing model.

Monetary incentives in crowdsourcing have been considered by \cite{ho2015incentivizing,ho2016adaptive,shah2016double}, and although the crowdsourcing setting is distinct from ours, these works help motivate our setting of paid stochastic experts.
\citet{ho2015incentivizing} empirically showed the effect of monetary incentives on the quality of the predictions in crowdsourcing. \citet{ho2016adaptive} introduce an online stochastic model where workers, who act strategically, are drawn i.i.d.\ from a fixed and unknown distribution of worker types. Each type determines the workers' productivity function and the workers' effort function, where the latter controls their strategic behavior. Because of strategic behaviors, their payment scheme is more complex than ours. On the other hand, in their model the requester's utility cannot be increased by aggregating workers on the same task.

Another example of strategic workers is investigated by \citet{shah2016double}, where they compute minimal payments sufficient to incentivize workers to predict labels that they are sure of, and abstain on the others.

\section{Preliminaries}\label{sec:prelim}

In each round $t \in [T]$ of online classification with paid experts, the learner chooses a payment $c_{t,j} \in [0,1]$ for each expert $j \in [K]$. After receiving the payments, the experts reveal their predictions $Z_{t,1}, \ldots, Z_{t,K} \in \{-1, 1\}^K$ for the true label $y_t \in \{-1,1\}$. For each $j \in [K]$, the prediction $Z_{t,j}$ is stochastic and satisfies \[
    \Pr(Z_{t,j} \neq y_t) = p_j(c_{t,j})
\]
where $p_j$ is the productivity function of expert $j$. Based on the expert advice, the learner then predicts $\yhatt \in \{-1, 1\}$, observes $y_t$, and suffers the zero-one loss $\id[\yhatt \not = y_t]$. The sequence $y_1,y_2,\ldots$ of true labels is arbitrary and deterministic\footnote{Our results continue to hold even when the labels are stochastic, provided that the events $\{Z_{t,j} \neq y_t\}$ remain independent.}, and the events $\{Z_{t,j} \neq y_t\}$ for $t \in [T]$ and $j \in [K]$ are stochastic and assumed to be independent. We let $Z_t=(Z_{t,1}, \dots, Z_{t,K})$ be the vector of all experts' predictions in round $t$.

We assume the experts' productivity functions $p_1,\ldots,p_K : [0,1] \to [0,1]$ are $L$-Lipschitz,
\begin{align}\label{eq:lipschitzassumption}
    \big|p_j(c) - p_j(c')\big| \leq L|c - c'| \qquad c,c' \in [0,1]~.
\end{align}
This class of productivity functions is broad enough to capture most realistic settings, including the special cases where the productivity funtion is monotonic, logistic, or where the experts are restricted to predict the correct label with probability greater than 0.5. 
These productivity functions are initially unknown to the learner. 
For any round $t=1,\dots, T$, we define the filtration
\[
    \Fset_t = \sigma(Z_1, B_1, y_1, \dots, Z_{t-1}, B_{t-1}, y_{t-1})
\]
where $B_s$ represents any internal randomization used by the learner in round $s\leq t$. The learner can use any information in $\Fset_t$ to estimate the productivity functions, decide on the payments and aggregate expert predictions in round $t$. The learners objective is to select payments and aggregate expert advice to minimize the cumulative regret defined in~\eqref{eqn:regdef}.

To understand the ideas behind the algorithm and the involved challenges, we first consider a simplified setting where payments do not affect the prediction accuracy. In other words, $p_j(c) = p'_j$ for all $c \in [0, 1]$, for all $j \in [K]$, and for some $p_1',\ldots,p_K' \in [0,1]^K$. %
As argued in the introduction, if $p'_1,\ldots,p'_K$ are known, then the learner's expected number of mistakes is exponentially decreasing in $K$. To see this,
let
$
    w(p') = \half \ln\frac{p'}{1-p'}
$
and assume the learner's prediction is
\begin{equation}
\label{eq:prediction}
    \yhatt = \sign\left(\sumK w(p_j')Z_{t,j}\right)~.
\end{equation}
Similarly to the analysis of AdaBoost \citep{schapire2013boosting}, we can upper bound the zero-one loss with the exponential loss and write
\begin{align}\label{eq:challenge1}
    &\Pr_{Z_t}(\yhatt \not = y_t) \leq \E_{Z_t}\left[\exp\left(-y_t\sumK w(p_j')Z_{t,j}\right)\right] \nonumber\\
    & = \prod_{j = 1}^K \left(p_j' \sqrt{\frac{1-p_j'}{p_j'}}  + (1-p_j')\sqrt{\frac{p_j'}{1-p_j'}}\right)\\ 
    & = \prod_{j = 1}^K \sqrt{4\big(1 - (\half - p_j')^2\big)} \leq \exp\left(-2\sumK (\half - p_j')^2\right)~. \nonumber
\end{align}
The first equality holds because of our choice of $w(p')$ and the assumption that the predictions of experts are independent. The last inequality uses $1 + x \leq e^x$. 

The tightness of the bound~\eqref{eq:challenge1}
is easily verified in the special case $(\half - p_j')^2 = \gamma^2$ for all $j \in [K]$. Then $\yhatt$ is the majority vote, which is clearly Bayes optimal and, assuming $K$ is odd,
\begin{align*}
    \Pr_Z(\yhatt \not = y_t)
&=
    \mathrm{Binom}\left(K, \left\lfloor\frac{K}{2}\right\rfloor, \frac{1}{2}+\gamma\right)
\\ &=
   \Omega\left(\sqrt{\frac{1}{K}}\,e^{\frac{K}{2}\ln(1-4\gamma^2)}\right)
\end{align*}
where $\mathrm{Binom}(n,m,p)$ is the probability of at most $m$ heads in $n$ flips of a coin with bias $p$, see \citet[inequality~(15)]{ferrante2021bounds}. This implies that, in the worst case, we are sub-optimal by a factor $K^{-1/2}$ as soon as we apply the first inequality in~\eqref{eq:challenge1}.

We now explain our approach for learning the unknown probabilities $p_j'$. Let $\phat_{t,j}$ be the estimate of $p_j'$ in round $t$. Following the derivation of~\eqref{eq:challenge1} and using prediction~\eqref{eq:prediction} with estimates $\phat_j$ in lieu of the true probabilities $p_j'$, we see that
\begin{align}\label{eq:challenge2}
    &\Pr_{Z_t}(\yhatt \not = y_t) \nonumber \\
    & \leq \prod_{j = 1}^K \left(p_j' \sqrt{\frac{1-\phat_{j}}{\phat_{j}}}  + (1-p_j')\sqrt{\frac{\phat_{j}}{1-\phat_{j}}}\right)
\end{align}
A first challenge is to control the difference between~\eqref{eq:challenge1} and~\eqref{eq:challenge2}. This involves controlling terms of order
\begin{align*}
    \sqrt{\big(1-\phat_{j}\big)\big/{\phat_{j}}} - \sqrt{\big(1-p_j'\big)\big/{p_j'}}~.
\end{align*}
Via standard online learning analysis, we would obtain a regret bound scaling linearly with the Lipschitz constant of the function $\sqrt{(1-p')/p'}$, which is of order $1/p'$. But this would require enforcing that $\min_j p_j'$ be bounded away from $0$ (and from $1$ for the symmetric function $\sqrt{p'/(1-p')}$).

A second challenge is learning the optimal cost for each expert. This is a bandit problem over continuously many actions, because choosing a payment $c \in [0,1]$ for some expert does not provide any information about payments $c' \neq c$. Using Lipschitzness of the productivity functions, we can discretize the payments. However, as we argued above, the key function $\sqrt{(1-p')/p'}$, which controls the error estimating the probability of mistake, is not Lipschitz in $[0,1]$. This necessitates further algorithmic developments and novel analyses.
In the following section we introduce our algorithm and explain how we overcome the aforementioned challenges.

\section{Algorithm}

\begin{algorithm}
\caption{LCB-GAPTRON}%
\label{alg:nosigmoid}
\begin{algorithmic}[1]
\REQUIRE{Set $\Cshat$ of $N$ costs, parameters $\beta \geq 0$ and $\delta > 0$}
\STATE{\textbf{Initialize:} For $j = 1, \ldots, K$ and $c \in \Cshat$, set $\hat{p}_{1, j}(c) = 1$, $\Phat_{1, j}(c) = 1$, $\alpha_{1, j}(c) = \half$, and $n_{1, j}(c) = 0$ \\
}
\FOR{$t = 1 \ldots T$}
\IF{$t \leq N$}
\STATE Find $c_{t, 1},\ldots,c_{t, K}$ such that $c_{t, k} = 0$ for all $k \in [K]$.
\ELSE
\STATE Compute $c_{t, 1},\ldots,c_{t, K}$ solution of
\begin{equation}
\label{eq:optimism}
\argmin_{c_1, \ldots, c_K \in \Cshat^K} 
    e^{- 2 \sumK \big(\half - \Phat_{t,j}(c_j)\big)^2} + \lambda \sumK c_j
\end{equation}
\ENDIF
\STATE Pay $c_{t, 1}, \ldots, c_{t, K}$ and receive advice $Z_{t, 1}, \ldots Z_{t, K}$
\IF{some $\hat{p}_{t, j}(c_{t,j}) \not \in \big[\alpha_{t, j}(c_{t,j}), 1-\alpha_{t, j}(c_{t,j})\big]$}
\STATE Pick any $j$ such that
\[
    \hat{p}_{t, j}(c_{t,j}) \not \in \big[\alpha_{t, j}(c_{t,j}), 1-\alpha_{t, j}(c_{t,j})\big]
\]
\STATE Predict $\yhatt = \sign(\hat{p}_{t, j}(c_{t,j}) - \half)Z_{t, j}$ 
\ELSE
\FOR{$j = 1,\ldots,K$}
\STATE $w_{t, j}(\phat_{t,j}(c_{t,j})) \leftarrow \half \log\frac{\phat_{t,j}(c_{t,j})}{1 - \phat_{t,j}(c_{t,j})}$
\ENDFOR
\STATE Set $x_t = \sumK w_{t, j}(\phat_{t,j}(c_{t,j})) Z_{t, j}$
\STATE Predict 
\begin{equation}\label{eq:randpredict}
    \yhatt = \begin{cases}
    \sign\left(x_t \right) & \textnormal{w.p. $1 - \half e^{-x_t\sign(x_t)}$}
    \\
    - \sign\left(x_t \right) & \textnormal{w.p. $\half e^{-x_t \sign(x_t)}$} 
    \end{cases}
\end{equation}
\ENDIF
\STATE Obtain true label $y_t$
\FOR{$j = 1, \ldots, K$}
\STATE $n_{t+1, j}(c_{t,j}) \leftarrow n_{t, j}(c_{t,j}) + 1$
\STATE $\alpha_{t+1, j}(c_{t,j}) \leftarrow \min\Big\{\frac{\beta}{n_{t+1, j}(c_{t,j})}, \half\Big\}$ 
\STATE $\phat_{t+1, j}(c_{t,j}) \leftarrow \sum_{s = 1}^{t} \id[c_{s,j} = c_{t,j}]\frac{1+Z_{s, j}y_t}{2 n_{t+1, j}(c_{t,j})}$
\STATE $s_{t+1}(c_{t,j}) \leftarrow \sign\big(\half - \phat_{t+1,j}(c_{t, j})\big)$
\STATE $Q_{t+1}(c_{t,j}) \leftarrow \min\bigg\{1 - \phat_{t+1, j}(c_{t,j}), \phat_{t+1, j}(c_{t,j}), $ $ \frac{3 \ln(3/\delta)}{n_{t+1,j}(c_{t, j})} +\sqrt{\frac{\phat_{t+1,j}(c_{t, j})(1-\phat_{t+1,j}(c_{t, j}))}{n_{t+1,j}(c_{t, j})}2\ln(\frac{3}{\delta}})\bigg\}$
\STATE $\Phat_{t+1, j}(c_{t,j}) \leftarrow \phat_{t+1,j} - s_{t+1}(c_{t,j})Q_{t+1}(c_{t,j})$ \label{line:optimistic} 
\ENDFOR
\ENDFOR
\end{algorithmic}
\end{algorithm}

Our algorithm LCB-GAPTRON for online classification with paid experts is presented in Algorithm~\ref{alg:nosigmoid}. At a high level, our algorithm selects payments using the pessimistic principle (since we receive bandit feedback for the payments), and then uses a randomized weighted aggregation procedure to predict a label based on the expert advice. However, as indicated above, several adjustments need to be made to account for the intricacies of the problem and ensure that we can obtain a $\scO(\sqrt{T})$ regret bound. We detail these below.

LCB-GAPTRON requires as input a discrete set $\Cshat$ of payments. To learn the optimal payment in $\Cshat$ for each expert, it is helpful to maintain an empirical estimate
\[
    \phat_{t+1, j}(c) = \sum_{s = 1}^{t} \id[c_{s,j} = c]\frac{1+Z_{s, j}y_t}{2 n_{t+1, j}(c)}
\]
of the probability of success for each $c \in \Cshat$, where $n_{t+1, j}(c) = \sum_{s = 1}^{t} \id[c_{s,j} = c]$ is the number of times we have paid expert $j$ $c$ up to the end of round $t$. We then construct optimistic estimates $\Phat_t(c)$ in line~\ref{line:optimistic} based on the empirical Bernstein bound \citep{audibert2007tuning}---see Lemma~\ref{lem:empbern} in the Appendix---which are used when computing the payments to the experts in~\eqref{eq:optimism}. 

The algorithm also requires a parameter $\beta \geq 0$, whose role is to control the cutoff value $\alpha_{t, j}(c)$ of the estimated probabilities for each $c \in \Cshat$. As discussed around~\eqref{eq:challenge2}, a key challenge in this problem is that the function that we optimize is subject to large changes when the estimated probabilities are close to $0$ or $1$. To solve this problem, when $p_{t,j}(c)$ is very large or very small, we simply follow experts $j$ that we estimate are either very good or very bad (in the latter case, the weight $w_{t,j}$ is negative). However, this does not resolve our troubles completely. Indeed, standard online methods would still suffer regret inversely proportional to the cutoff value because they need to control the difference between the estimated probabilities and the true probabilities. To overcome this issue, we show that we can estimate the true probabilities up to a multiplicative factor of $\tfrac{3}{2}$. This is also a result of using the empirical Bernstein bound to construct the confidence intervals. Indeed, the empirical Bernstein bound allows us to have both additive and multiplicative control of the estimated probabilities which is essential for our analysis. See Section~\ref{sec:analysis} for further details.  

To avoid suffering additional regret for only being able to estimate the probabilities up to a multiplicative factor, we use the ideas of \citet{VanderHoeven2020exploiting, vanderHoeven2021beyond}. In particular, whenever all the estimated probabilities are bounded away from $0$ and $1$, we use randomized predictions. These randomized predictions $\yhatt = \yhatt(B_t,Z_t)$ defined in~\eqref{eq:randpredict}---where $B_t$ is the internal randomization used by Algorithm~\ref{alg:nosigmoid} at step $t$---satisfy (see Lemma~\ref{lem:gaptronbound})
\begin{align*}%
    \Pr_{B_t}\big(\yhatt \not = y_t \mid Z_t, \Fset_t\big)
    \leq \half e^{ - y_t \sumK w_{t, j}(\phat_{t,j}(c_{t,j}))Z_{t,j}}
\end{align*}
where $c_{t,j} \in \Cshat$ is the payment to expert $j$ in round $t$ and  $\phat_{t,j}(c_{t,j})$ is the estimate of the accuracy of expert $j$ when the payment is $c_{t,j}$. 
This gains us a factor $\half$ compared to the bound we used in Section~\ref{sec:prelim} (see equation~\eqref{eq:challenge1}) and allows us to compensate for the multiplicative factor introduced by estimating the probabilities. %

\section{Analysis}\label{sec:analysis}
In this section we prove the regret bounds in the introduction. To condense notation slightly, let 
\begin{align*}
    \Phi\big(\p(\c)\big) = e^{-2\sumK\big(\frac{1}{2} - p_j(c_j)\big)^2}
\end{align*}
where $\p(\c)$ is the vector with elements $p_j(c_j)$. Then, $\OPT(\p) = \min_{\c\in [0,1]^K} \Phi\big(\p(\c)\big) + \lambda\sumK c_j$. Our proof of the regret bound follows from the below steps. 

{\bf Step 1: Bound the error in estimated probabilities}
We start by showing that we can control the difference between true probabilities $p_j(c)$ and estimated probabilities $\phat_{t,j}(c)$. For any $\delta \in (0,1)$ and $t \ge 1$, let $\Lambda_{\delta, t}$ be the event that
    \begin{align}\label{eq:lambdadef}
        & \big|\phat_{t, j}(c) - p_{j}(c)\big| \leq \frac{3}{n_{t, j}(c)}\ln\frac{3}{\delta} \nonumber \\
        & + \sqrt{\frac{\phat_{t, j}(c)(1-\phat_{t, j}(c))}{n_{t, j}(c)}2\ln\frac{3}{\delta}}  
    \end{align} 
    simultaneously for all $j \in [K]$ and $c \in \Cshat$. 
\begin{restatable}{lemma}{lemdifcontrol}
\label{lem:difcontrol}
    Fix $\delta \in (0,1)$. Suppose that $t > N$. Then $\Lambda_{\delta, t}$ holds with probability at least $1 - \delta TNK$.
\end{restatable}
The proof of Lemma~\ref{lem:difcontrol} follows from an application of the union bound and the empirical Bernstein bound \citep{audibert2007tuning} and is provided in the Appendix. 

{\bf Step 2: Bound the prediction error}
We now turn our attention to controlling the number of mistakes we make. In the following, we bound the probability that we make a mistake in any given round $t$. We split the analysis into two cases: either all $\hat{p}_{t, j}(c_{t,j}) \in \big[\alpha_{t, j}, 1 - \alpha_{t, j}(c_{t,j})\big]$, or at least one $\hat{p}_{t, j}(c_{t,j})$ is outside of the specified range (meaning that some expert is certain so we can predict just using their advice). We denote by $\Epsi_{t}$ the event that in round $t$ we are in the first case, that is that $\hat{p}_{t, j}(c_{t,j}) \in \big[\alpha_{t, j}(c_{t,j}), 1-\alpha_{t, j}(c_{t,j})\big]$ for all $j$.

As a first step, observe that when $\Epsi_t$ holds we issue a randomized prediction, see equation~\eqref{eq:randpredict} in Algorithm~\ref{alg:nosigmoid}. This prediction has the following crucial property. 
\begin{restatable}{lemma}{lemgaptronbound}
\label{lem:gaptronbound}
    For any round $t > N$, assuming $\Epsi_t$ holds, the prediction in equation \eqref{eq:randpredict} satisfies
    \begin{align*}
        P_{B_t}(\yhatt \not = y_t|\Fset_{t}, Z_t) \leq \half e^{-y_t \sumK w_{t, j}(\phat_{t,j}(c_{t,j}))Z_{t,j}}.
    \end{align*}
\end{restatable}
The proof of Lemma~\ref{lem:gaptronbound} can be found in the appendix and essentially follows from \citet[Lemma~1]{vanderHoeven2021beyond} although our proof is simpler. 

As a second step, let us integrate out the randomness in $Z_t$ in the bound in Lemma~\ref{lem:gaptronbound}. Using the definition of $w_{t,j}$ we obtain:
\begin{align}\label{eq:analysissqrtphat}
     P_{B_t,Z_t}(\yhatt & \not = y_t|\Fset_{t}) \leq 
    \frac{1}{2} \prod_{j = 1}^K \Bigg(p_{j}(c_{t,j}) \sqrt{\frac{1-\phat_{t, j}(c_{t,j})}{\phat_{t, j}(c_{t,j})}}  \nonumber \\
    & + (1-p_j(c_{t,j}))\sqrt{\frac{\phat_{t, j}(c_{t,j})}{1-\phat_{t, j}(c_{t,j})}}\Bigg).
\end{align}
Now, under the assumptions that all $\hat{p}_{t, j}(c_{t,j}) \in \big[\alpha_{t, j}(c_{t,j}), 1-\alpha_{t, j}(c_{t,j})\big]$ and that $\Lambda_{\delta, t}$ holds, for all $\beta > 0$ 
$$
\big|\phat_{t,j}(c_{t,j}) - p_j(c_{t,j})\big| = \tilde{\scO} \left(\frac{1}{\beta} \phat_{t,j}(c_{t,j})\right)~.
$$ 
To see why this holds, observe that $\hat{p}_{t, j}(c_{t,j}) \geq \alpha_{t, j}(c_{t,j}) = \scO\big(\frac{\beta}{n_{t, j}(c_{t,j})}\big)$, which we can use to bound the right-hand side of the equation in the definition of $\Lambda_{\delta, t}$ in equation~\eqref{eq:lambdadef}. %
We can then substitute this into equation~\eqref{eq:analysissqrtphat} and, after some manipulation together with a careful choice of $\beta$, show that when $\Epsi_t$ and $\Lambda_{\delta,t}$ both hold,
\begin{align*}
    \Pr_{B_t,Z_t}(\yhatt \not = y_t|\Fset_t)
    \leq \frac{3}{4}\prod_{j = 1}^K \sqrt{1 - 4\big(\half-\phat_{t,j}(c_{t,j})\big)^2}
\end{align*}
after which, using $1 + x \leq \exp(x)$, we arrive at the statement of Lemma~\ref{lem:multiplicativebound1}. 
\begin{restatable}{lemma}{lemmultiplicativeboundone}
\label{lem:multiplicativebound1}
    Let $\beta = {18\ln(3/\delta)K^2}$. Then for any round $t > N$ and $\delta \in (0, 1)$, assuming $\Lambda_{\delta, t}$ and $\Epsi_t$ both hold,
    \begin{align*}%
        & \Pr_{B_t,Z_t}(\yhatt \not = y_t|\Fset_t) \leq \frac{3}{4} \Phi\big(\bphat_{t}(\c_{t})\big)%
    \end{align*}
\end{restatable}
{\bf Step 3: Relate  $\Phi\big(\hat\p(\c)\big)$ to $ \Phi\big(\p(\c)\big)$}
In the next lemma we show how to control the difference between the right-hand side of Lemma~\ref{lem:multiplicativebound1} and the same equation, but with $p_j(c_{t,j})$ instead of $\phat_{t,j}(c_{t,j})$.
\begin{restatable}{lemma}{lemmultiplicativeboundtwo} %
\label{lem:multiplicativebound2}
     For any round $t > N$ and $\delta \in (0, 1)$, assuming $\Lambda_{\delta, t}$ holds,
    \begin{align}
        & \frac{3}{4} \Phi\big(\bphat_t(\c_t)\big) %
        \leq \frac{7}{8} \Phi\big(\p(\c_t)\big) \nonumber \\ %
        & \quad + 96 K \sumK \left(\frac{2\ln(3/\delta)}{n_{t,j}(c_{t,j})} + \left(\frac{3\ln(3/\delta)}{n_{t,j}(c_{t,j})}\right)^2\right). \nonumber 
    \end{align}
\end{restatable}
To prove Lemma~\ref{lem:multiplicativebound2} (see the Appendix for a detailed proof) we show that for some constant $b >0$ and for all $\q,\q' \in [0,1]^K$,
\begin{align*}
    \Phi(\q) - \Phi(\q') & \leq \tfrac{1}{8} \Phi(\q) +  b K\sumK |q_j - q'_j|^2~.
\end{align*}
This means that 
\begin{align*}%
    \tfrac{3}{4} \Phi(\q) & = \tfrac{7}{8} \Phi(\q) - \tfrac{1}{8} \Phi(\q) \nonumber \\
    & \leq \frac{7}{8} \left(\Phi(\q') + b K\sumK |q_j - q'_j|^2\right)~.
\end{align*}
Since we assumed that the event $\Lambda_{\delta, t}$ holds, we can replace $q_j$ and $q_j'$ by $\phat_{t,j}(c_{t,j})$ and $p_j(c_{t,j})$ to prove Lemma~\ref{lem:multiplicativebound2}. 

{\bf Step 4: Bound the loss from pessimistic choice of payments}
Next, we need to control the difference between paying the experts our chosen costs versus the optimal costs. Using the optimism in the face of uncertainty principle, the same ideas we used to prove Lemma~\ref{lem:multiplicativebound2}, and the Lipschitz assumption in equation~\eqref{eq:lipschitzassumption} we arrive at the following lemma, whose proof can be found in the Appendix.
\begin{restatable}{lemma}{lemoptimism} 
\label{lem:optimism}
    Let $\Cshat$ be such that for any $c^\star \in [0,1]$ there is a $\tilde{c} \in \Cshat$ that satisfies $|\tilde{c} - c^\star| \leq \varepsilon$. Then for any round $t > N$ and $\delta \in (0, 1)$, assuming $\Lambda_{\delta, t}$ holds,
    \begin{align*}
        & \frac{7}{8} \Phi\big(\p(\c_t)\big) + \lambda \sumK c_{t,j} \\
        & \leq \min_{\c \in [0,1]^K}\Bigg\{ \Phi\big(\p(\c)\big) + \lambda \sumK c_{j} \Bigg\} + (4L + \lambda)K\varepsilon \\ 
        & \quad + 96 K\sumK \left(\frac{2\ln(3/\delta)}{n_{t,j}(c_{t,j})} 
        + \left(\frac{3\ln(3/\delta)}{n_{t,j}(c_{t,j})}\right)^2\right)~.
    \end{align*}
\end{restatable}
{\bf Step 5: Control the regret in when estimated probabilities are too large or small}
Combined, Lemmas~\ref{lem:difcontrol}, \ref{lem:multiplicativebound1}, and  \ref{lem:multiplicativebound2} give us control over the regret in rounds where all $\hat{p}_{t, j}(c_{t,j}) \in \big[\alpha_{t, j}(c_{t,j}), 1-\alpha_{t, j}(c_{t,j})\big]$. In the case where there is at least one $\hat{p}_{t, j}(c_{t,j})$ which is not in the above range, then we can control the regret by using the following observation. In rounds where $\Epsi_t$ does not hold we follow the (flipped) prediction of an expert $j$ that satisfies  $\hat{p}_{t, j}(c_{t,j}) \not \in \big[\alpha_{t, j}(c_{t,j}), 1-\alpha_{t, j}(c_{t,j})\big]$. For simplicity suppose that we just follow the actual prediction of that expert, which implies that $\hat{p}_{t, j}(c_{t,j}) \leq \alpha_{t, j}(c_{t,j})$. In this case $\yhatt$ does not depend on $B_t$ and we have that, assuming $\Lambda_{\delta, t}$ and $\Epsi_t^{c}$ both hold,%
\begin{align*}
    \Pr_{Z_t}(\yhatt \not = & y_t \mid \Fset_{t}) = p_j(c_{t,j}) \\
    & \leq \phat_{t, j}(c_{t,j}) + |p_j(c_{t,j}) - \phat_{t, j}(c_{t,j})| \\
    & = \tilde{\scO}(\alpha_{t,j}(c_{t,j})) = \tilde{\scO}\Big(\frac{\beta}{n_{t, j}(c_{t,j})}\Big)~,
\end{align*}
where the third equality follows from the assumption that events $\Lambda_{\delta, t}$ and $\Epsi_t^{c}$ hold. To see why, observe that since $\phat_{t, j}(c_{t,j}) \leq \alpha_{t, j}(c_{t,j})$ equation~\eqref{eq:lambdadef} is of order $\alpha_{t, j}(c_{t,j})$. The formal result can be found in Lemma~\ref{lem:notinrange} below, the full proof of which can be found in the Appendix. 
\begin{restatable}{lemma}{lemnotinrange} 
\label{lem:notinrange}
    For any round $t > N$ and $\delta \in (0, 1)$, assuming $\Epsi_t^c$ and $\Lambda_{\delta,t}$ both hold,
    \begin{align*}
        & \Pr_{Z_t}\big(\yhatt \not = y_t \mid \Fset_t \big) + \lambda \sumK c_{t, j}  \\
        & \leq \OPT(\p) 
        + \frac{2\beta + 4\ln(3/\delta)}{n_{t, j_t}(c_{t,j_t})} + (4L + \lambda)K\varepsilon  \\
        & \quad + 96 K\sumK \Bigg(\frac{2\ln(3/\delta)}{n_{t,j}(c_{t,j})} 
         + \Big(\frac{3\ln(3/\delta)}{n_{t,j}(c_{t,j})}\Big)^2\Bigg).
    \end{align*}
\end{restatable}
{\bf Step 6: Sum up the per-round regret}
We now have a per-round control of the cost and probability of making a mistake. Thus, by combining Lemmas~\ref{lem:difcontrol}, \ref{lem:multiplicativebound1}, \ref{lem:multiplicativebound2}, \ref{lem:optimism}, and \ref{lem:notinrange} and summing over rounds we arrive at the main result of the paper in Theorem~\ref{th:mainO} below, whose result is implied by Theorem~\ref{th:maindetail} in the Appendix. As stated below, the choice of $\Cshat$ in Algorithm~\ref{alg:nosigmoid} that delivers the bounds of Theorem~\ref{th:mainO} is the uniform $\varepsilon$-grid on $[0,1]$.
\begin{restatable}{theorem}{maintheoremO} 
\label{th:mainO}
    Let $\Cshat$ be such that for any $c^\star \in [0,1]$ there is a $\tilde{c} \in \Cshat$ that satisfies $|\tilde{c} - c^\star| \leq \varepsilon$, let $\beta = {18\ln(3/\delta)K^2}$, and let $\delta = \frac{1}{(1 + \lambda K)T^2K}$. Then the regret of Algorithm~\ref{alg:nosigmoid} satisfies
    \begin{align*}
    R_T = \scO\left( \varepsilon KT (4L + \lambda K) + N \Big(\lambda  + K^3 \ln(\lambda K T)^2  \Big)  \right)~.
    \end{align*}
\end{restatable}
Theorem~\ref{th:mainO} implies that if $N = \scO(\varepsilon^{-1})$ for $\varepsilon$ of order $\sqrt{\big(\lambda  + K^3 \ln(\lambda K T)^2  \big)\big/\big(KT (4L + \lambda K)\big)}$,
then our final bound is 
\begin{align}\label{eq:mainbigOtuned}
R_T = \scO\left( K^2\sqrt{T (4L + \lambda)\Big(\lambda  + \ln(1\lambda K T)^2  \Big)}  \right).
\end{align}
A slight modification of the proof shows that if we compete with the best costs in $\Cshat$, rather than the best costs in $[0,1]^K$, then the corresponding regret,
\begin{align*}
    R_T(\Cshat)
&=
    \E\left[\sumT \Big(\id[\yhatt \not = y_t] + \lambda \sumK c_{t, j}\Big)\right] \\
&\quad -
    T \min_{\c \in \Cshat}\Bigg\{\Phi\big(\p(\c)\big) + \lambda \sumK c_{j} \Bigg\}
\end{align*}
can be bounded by $\scO\Big(N \big(\lambda  + K^3 \ln(\lambda K T)^2  \big)\Big)$. Note that in this case, we remove the first term of the regret in Theorem~\ref{th:mainO} as there is no discretization error. Moreover, if we have a discrete set of costs $\Cshat$, then we do not require the costs to be Lipschitz.

\section{Alternative Approaches}
\label{sec:alternative}
In this section we formulate alternative notions of regret that make the problem of online classification with paid experts amenable to solution via standard bandit algorithms that predict using the advice of a single expert. We start by considering a finite set $\Cshat$ of costs and assume that in each round the learner must pick a single expert and a cost $c \in \Cshat$ to pay them. Assuming $p_j(c) \ge \frac{1}{2}$ for all experts $j \in [K]$ and costs $c \in \Cshat$, we may define the regret
\begin{align*}
    R^{\mathrm{band}}_T &= \E\left[\sumT \Big(\id[\yhatt \not = y_t] + \lambda c_{t}\Big)\right]
\\&\quad
    - T \min_{j \in [K], c \in \Cshat}\big(p_j(c) + \lambda c \big)~.
\end{align*}
$R^{\mathrm{band}}_T$ presents a different trade-off between costs and mistakes. In particular, while the term accounting for the costs is considerably smaller (as only one expert is paid in each round), the term accounting for the expected number of mistakes is exponentially larger (in the number of experts) because there is no expert aggregation. Treating each expert and cost pair as an arm, we can use LCB (the variant of the UCB algorithm using lower confidence bounds to minimize losses instead of maximizing rewards) and immediately obtain the bound $R^{\mathrm{band}}_T = \tilde{\scO}\big(\sqrt{K N T}\big)$.

When the productivity functions are $L$-Lipschitz, see~\eqref{eq:lipschitzassumption}, we can define a harder notion of regret $R^{\mathrm{cont}}_T$ in which the comparator is defined with respect to the best cost $c \in [0,1]$, as opposed to the best $c\in\Cshat$ which we used in the definition of $R^{\mathrm{band}}_T$.
Now, if we discretize the interval $[0,1]$ using the grid $\Cshat$ of Theorem~\ref{th:mainO} with $|\Cshat| = \scO\big(\varepsilon^{-1}\big)$, then we pay an approximation error of $\varepsilon T(L+\lambda)$. By running LCB on $\Cshat\times K$ arms, and adding the approximation error, we get, after tuning $\varepsilon$,
$
    R^{\mathrm{cont}}_T %
    = \scO\big(T^{2/3}(KL)^{1/3}\big)
$.

Although these bounds on $R^{\mathrm{band}}_T$ and $R^{\mathrm{cont}}_T$ are not directly comparable to the bounds on our different notion of regret~\eqref{eqn:regdef}, in the next section we perform an empirical comparison between variants of Algorithm~\ref{alg:nosigmoid} and the instance of LCB run on $K \times N$ arms which we used to bound $R^{\mathrm{band}}_T$.

\section{Experiments}\label{sec:experiments}
Our experiments use two sets of productivity functions defined on a uniform (random) grid $\Cshat$ of $N$ payments on $\big[\frac{1}{2},1\big]$.
The first productivity function is linear with the same slope for all experts: We sampled $N$ numbers $c_1,\ldots,c_N$ uniformly at random from $\big[\frac{1}{2},1\big]$ and defined $p_j(c_i) = c_i$ for all $i \in [N]$ and all $j \in [K]$.
The second productivity function is sigmoidal, with a different slope for each expert: We sampled $N$ numbers $c_1,\ldots,c_N$ uniformly at random from $[0,1]$ and $K$ integers $\theta_1,\ldots,\theta_K$ uniformly at random from $[1,10]$. For each $j \in [K]$ we then set
\begin{equation}
    \label{eq:prod2}
    p_j(c_i) = \frac{\exp(\theta_j c_i)}{1+\exp(\theta_j c_i)}~.
\end{equation}
A fresh sample of the productivity functions for each expert is drawn in each repetition of our experiments.

Consistent with our definition of regret, we measure the performance of the algorithms in terms of their cost; i.e., the total number of mistakes plus the total payments to the experts. 

The running time for finding the optimal costs in $T$ rounds using our optimistic estimates~\eqref{eq:optimism} in LCB-GAPTRON (Algorithm~\ref{alg:nosigmoid}) is $\scO(TN^K)$, which prevents running experiments for moderately large values of the parameters. Therefore, we designed two simple and efficient approximations of the optimal payments defined in~\eqref{eq:optimism}. The first one, \textsc{selfish}, optimizes the cost for each expert independently of the others. The second one, \textsc{local}, computes the optimal cost of each expert iteratively, in a round-robin fashion, while keeping the cost of the other experts fixed. We call \textsc{brute} the inefficient implementation of LCB-GAPTRON that directly optimizes~\eqref{eq:optimism} using brute force search. Finally, we call LCB the instance of LCB run over $K \times N$ actions which we defined in Section~\ref{sec:alternative}.

\begin{figure}
    \centering
    \begin{subfigure}
         \centering
         \includegraphics[width=0.35\textwidth]{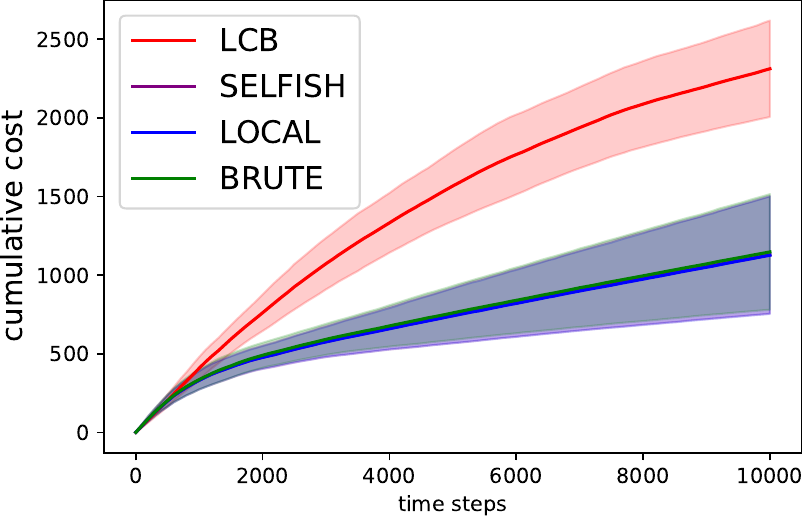}
     \end{subfigure}
     \begin{subfigure}
         \centering
         \includegraphics[width=0.35\textwidth]{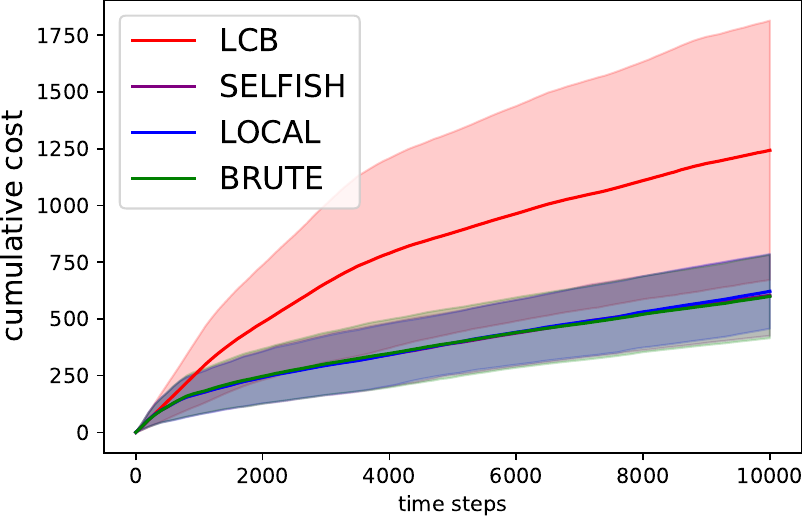}
     \end{subfigure}
    \caption{Cumulative cost over time
    for the choice of parameters $K=5$, $N=5$, $T=10^4$, $\lambda = 10^{-2}$.
    The algorithms are run using the first productivity function in the top plot and the second productivity function in the bottom plot. The error bars show the standard deviation of the cost averaged over $20$ repetitions.}
    \label{fig: small_exps}
\end{figure}

In our first experiment we pick
sufficiently small values for the parameters so that we can run LCB-GAPTRON-\textsc{brute}. Figure~\ref{fig: small_exps} shows that for both choices of the productivity function, all instances of LCB-GAPTRON have nearly indistinguishable performance and outperform LCB. Thus, we can safely drop LCB-GAPTRON-\textsc{brute} and run a second set of experiments using a larger time horizon $T=10^5$.

 \begin{figure}[h]
    \centering
    \begin{subfigure}
         \centering
         \includegraphics[width=0.35\textwidth]{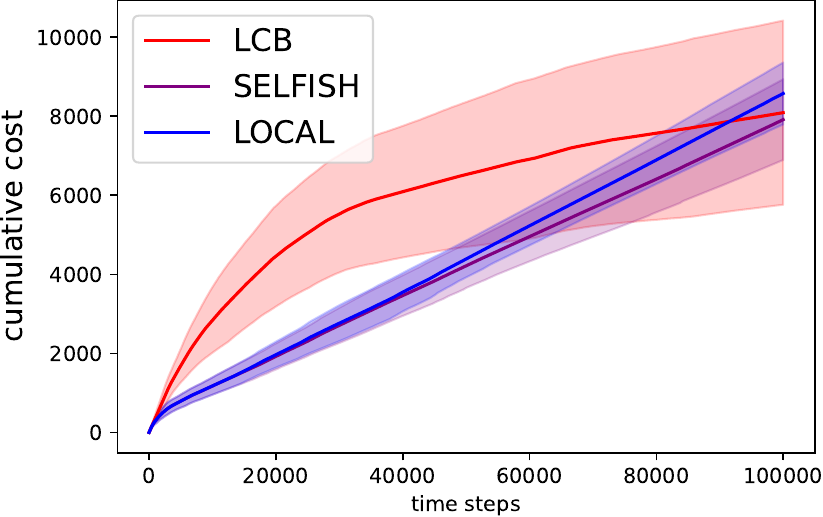}
     \end{subfigure}
     \begin{subfigure}
         \centering
         \includegraphics[width=0.35\textwidth]{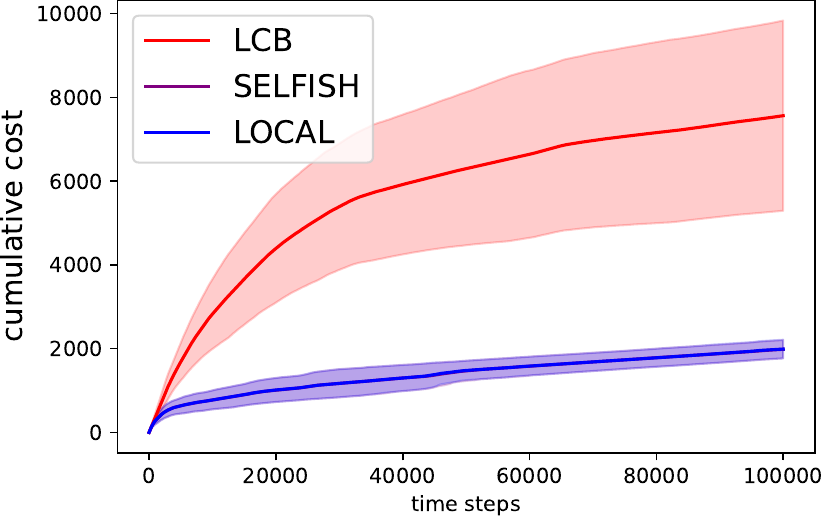}
     \end{subfigure}
        \begin{subfigure}
         \centering
         \includegraphics[width=0.35\textwidth]{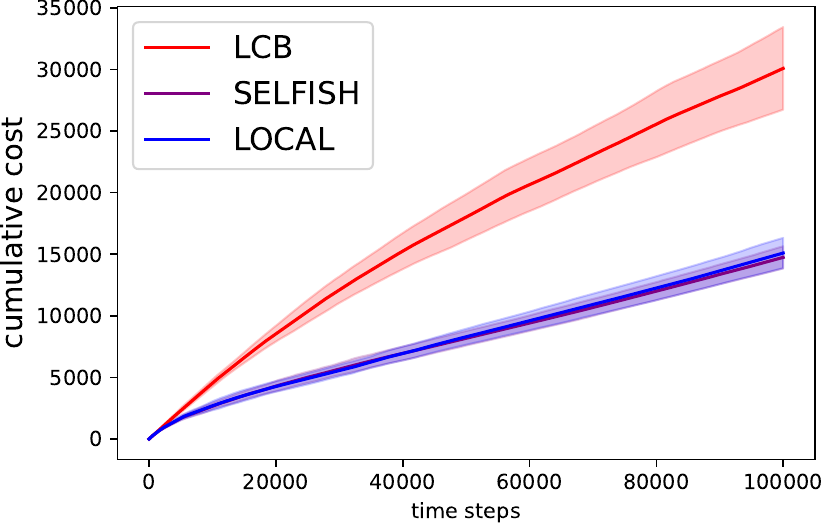}
     \end{subfigure}
    \caption{Total cost for the following choices of parameters: $K=10,N=10,T=10^5,\lambda=10^{-2}$ (top plot), $K=10,N=10,T=10^5,\lambda=10^{-3}$ (central plot), $K=20,N=50,T=10^5,\lambda=10^{-2}$ (bottom plot). The algorithms are run using the second productivity function. The error bars show the standard deviation of the cost averaged over $20$ repetitions.}
    \label{fig: large_exps}
\end{figure}

In our second experiment, we test LCB-GAPTRON-\textsc{selfish} and LCB-GAPTRON-\textsc{local} against LCB using the second productivity function (plots using the first productivity function are similar, see Appendix~\ref{app:experiments}). The results in Figure~\ref{fig: large_exps} show that for small value of the scaling parameters ($K=10$ and $N=10$) and large cost units ($\lambda = 10^{-2}$), LCB is eventually on par with LCB-GAPTRON. Recall that LCB pays a single expert at each round, which is the reason that LCB catches up with LCB-GAPTRON. However, for larger values of the scaling parameters ($K=20$ and $N=50$) or for smaller values of the cost units ($\lambda = 10^{-3}$), both variants of LCB-GAPTRON dominate again.

In Appendix~\ref{app:experiments}, we report the performance of LCB, LCB-GAPTRON-\textsc{selfish}, and LCB-GAPTRON-\textsc{local} on both productivity functions for $T = 10^5$, $K \in \{10,20\}$, $N \in \{10,50,100\}$, and $\lambda \in \big\{10^{-2},10^{-3}\big\}$. These plots essentially confirm the observations made in the discussion of Figure~\ref{fig: large_exps}. In Appendix~\ref{app:experiments} we also provide the figures for the experiments with LCB-GAPTRON-\textsc{brute} with $\lambda \in \big\{10^{-2},10^{-3}\big\}$, $N = 5$, $K = 5$, $T = 10^5$, and both productivity functions, which also confirm the observations made in the discussion of Figure~\ref{fig: small_exps}.

\section{Future Work}
In this paper we have studied online classification with paid stochastic experts, and presented an algorithm which has sub-linear regret in terms of the number of rounds $T$. We have also demonstrated that this algorithm performs well in several experimental settings. Although the algorithm is computationally expensive, we have shown empirically that using approximations does not lead to much deterioration of performance. Several questions remain open. For example, whether a computationally efficient algorithm with similar regret guarantees can be developed, and whether the $K^2$ factor in our regret bound $K^2(\ln T)\sqrt{T}$ can be improved on. On the other hand, we conjecture the $\sqrt{T}$ rate is optimal, because of the need of estimating the discretized productivity functions from bandit feedback.

There is scope to extend our model in several directions by considering, for example, strategic experts (as in \citet{roughgarden2017online, freeman2020no}), or making the experts' performance to depend also on contextual information. It is also an open question whether faster rates can be achieved with stronger parametric assumptions on the productivity function. For example, if the productivity function is a sigmoid, $p_j(c) = \exp(a+bc)/(1+\exp(a+bc))$ with unknown constants $a,b, \in \mathbb{R}$, can the regret be significantly improved?

\section*{Acknowledgements}
This work was mostly done while DvdH was at the University of Milan partially supported by the MIUR PRIN grant Algorithms, Games, and Digital Markets (ALGADIMAR) and partially supported by Netherlands Organization for Scientific Research (NWO), grant number VI.Vidi.192.095. HQ and NCB are partially supported by the MIUR PRIN grant Algorithms, Games, and Digital Markets (ALGADIMAR), by the EU Horizon 2020 ICT-48 research and innovation action under grant agreement 951847, project ELISE (European Learning and Intelligent Systems Excellence), and by the FAIR (Future Artificial Intelligence Research) project, funded by the NextGenerationEU program within the PNRR-PE-AI scheme (M4C2, investment 1.3, line on Artificial Intelligence).
CPB gratefully acknowledges the support of the Imperial College European Partners Fund.

\bibliography{expertrefs,ncb}
\bibliographystyle{apalike}

\newpage
\appendix
\onecolumn

\section{PROOF DETAILS}\label{app:proofdetials}
\begin{lemma}[\citet{audibert2007tuning}]\label{lem:empbern}
    Let $X_1, \ldots, X_t$ be i.i.d.\ random variables with $\E[X] = \mu$ taking their values in $[0, 1]$. Denote by $\bar{X}_t = \frac{1}{t}\sum_{s=1}^t X_s$ and by $V_t = \frac{1}{t} \sum_{s=1}^t (X_s - \bar{X}_t)^2$. Then with probability at least $1 - \delta$
    \begin{align*}
        |\bar{X}_t - \mu| \leq \sqrt{\frac{V_t}{t}2\ln(3/\delta)} + \frac{3\ln(3/\delta)}{t}.
    \end{align*}
\end{lemma}

\lemgaptronbound*

\begin{proof}
We denote by $x_t = \sumK w_{t, j}(\phat_{t,j}(c_{t,j})) Z_{t, j}$
The proof also follows from Lemma~1 by \citet{vanderHoeven2021beyond}. Suppose that $\sign(x_t) = \sign(y_t)$. Then $\E[\id[\yhatt \not = y_t]|\Fset_{t}, Z_t] = \half \exp(-y_t x_t)$. Now suppose that $\sign(x_t) \not = \sign(y_t)$. Then 
\begin{align*}
    \E_{\yhatt}[\id[\yhatt \not = y_t]|\Fset_{t}, Z_t] & = 1 - \half \exp(y_t x_t) = 1 + \half \exp(-y_t x_t) - \half \exp(y_t x_t) - \half \exp(- y_t x_t) \leq \half \exp(-y_t x_t),
\end{align*}
where the last inequality is due to $\half \exp(y_t x_t) + \half \exp(- y_t x_t) \geq 1$, which holds by Jensen's inequality.
\end{proof}

\lemdifcontrol*

\begin{proof}

First, note that since $t>N$ $n_t(c) \geq 1$ for all $c \in \Cshat$. 
Denote by $V_{t, j}(c) = \frac{1}{|\mathcal{N}_t|}\sum_{i \in \mathcal{N}_t}\big(\sum_{i' \in \mathcal{N}_t}\frac{1 + Z_{i', j}y_t}{2|\mathcal{N}_t|} - \frac{1+Z_{i, j}y_t}{2}\big)^2$, where $\mathcal{N}_t = \{i<t: \id[c_{i, j} = c]\}$ and denote by 
\begin{align*}
    \hat{V}_{t, j}(c) & = \frac{1}{n_{t,j}(c)}\sum_{s = 1}^{t-1}\id[c_{s, j} = c]\Bigg(\sum_{s' = 1}^{t-1}\id[c_{s', j} = c]\frac{1+Z_{s', j}y_t}{n_{t,j}(c)} - \frac{1+Z_{s, j}(c_{s,j})y_t}{2}\Bigg)^2 \\
    & = \phat_{t,j}(c)(1-\phat_{t,j}(c))
\end{align*}
Now, by reindexing the sum we can see that
\begin{align*}
    & \frac{1}{n_{t,j}(c)}\sum_{s = 1}^{t-1}\id[c_{s, j} = c]\Bigg(\sum_{s' = 1}^{t-1}\id[c_{s', j} = c]\frac{1+Z_{s', j}y_t}{n_{t,j}(c)} - \frac{1+Z_{s, j}(c_{s,j})y_t}{2}\Bigg)^2 \\
    & = \frac{1}{|\mathcal{N}_t|}\sum_{i \in \mathcal{N}_t}\big(\sum_{i' \in \mathcal{N}_t}\frac{1 + Z_{i', j}y_t}{2|\mathcal{N}_t|} - \frac{1+Z_{i, j}y_t}{2}\big)^2.
\end{align*}
Therefore, we can see that 
\begin{align*}
    & P\Bigg(|\phat_{t, j}(c) - p_{j}(c)| \geq \sqrt{\frac{\hat{V}_{t, j}(c)}{n_{t, j}(c)}2\ln(3/\delta)} + \frac{3\ln(3/\delta)}{n_{t, j}(c)}\Bigg) \\
    & \leq P\Bigg(\exists~\mathcal{N}_h \in \{\mathcal{N}_{N+1}, \ldots, \mathcal{N}_t\}  \quad s.t. \quad |\sum_{i \in \mathcal{N}_h}\frac{1 + Z_{i, j}(c)y_t}{2|\mathcal{N}_h|} - p_{j}(c)| \geq \sqrt{\frac{{V}_{h, j}(c)}{|\mathcal{N}_h|}2\ln(3/\delta)} + \frac{3\ln(3/\delta)}{|\mathcal{N}_h|}\Bigg) \\
    & \leq \sum_{h = N+1}^t P\Bigg(|\sum_{i \in \mathcal{N}_h}\frac{1 + Z_{i, j}(c)y_t}{2|\mathcal{N}_h|} - p_{j}(c)| \geq \sqrt{\frac{V_{n, j}(c)}{|\mathcal{N}_h|}2\ln(3/\delta)} + \frac{3\ln(3/\delta)}{|\mathcal{N}_h|}\Bigg) \leq T \delta,
\end{align*}
where the last inequality is due to Lemma~\ref{lem:empbern}. Taking the union bound we can see that the above holds for all $j$ and $c \in \Cshat$ with probability at most $TNK\delta$. 
\end{proof}

\lemmultiplicativeboundone*

\begin{proof}
Since both $\Lambda_{\delta,t}$ and $\Epsi_t$ hold we may use Lemma~\ref{lem:gaptronbound} to obtain 
\begin{align*}
    & \E_{Z}[\E_{\yhatt}[\id[\yhatt \not = y_t]]]  \leq \half \E_{Z}[\exp(-y_t x_t)]\\
    & = \half\prod_{j = 1}^K \left(P(Z_{t, j}(c_{t,j})y_t = 1)\sqrt{\frac{1 - \phat_{t,j}(c_{t,j})}{\phat_{t,j}(c_{t,j})}}  + P(Z_{t, j}(c_{t,j})y_t = -1)\sqrt{\frac{\phat_{t,j}(c_{t,j})}{1 - \phat_{t,j}(c_{t,j})}}\right) \\
    & = \half\prod_{j = 1}^K \left(p_{j}(c_{t,j})\sqrt{\frac{1 - \phat_{t,j}(c_{t,j})}{\phat_{t,j}(c_{t,j})}}  + (1 - p_{j}(c_{t,j}))\sqrt{\frac{\phat_{t,j}(c_{t,j})}{1 - \phat_{t,j}(c_{t,j})}}\right) \\
    & = \half\prod_{j = 1}^K \Bigg(2\sqrt{{\phat_{t,j}(c_{t,j})}({1 - \phat_{t,j}(c_{t,j})})}\\
    & \quad + \big(p_{j}(c_{t,j}) - \phat_{t, j}(c_{t,j})\big)\sqrt{\frac{1 - \phat_{t,j}(c_{t,j})}{\phat_{t,j}(c_{t,j})}}  + \big(\phat_{t, j}(c_{t,j}) - p_{j}(c_{t,j})\big)\sqrt{\frac{\phat_{t,j}(c_{t,j})}{1 - \phat_{t,j}(c_{t,j})}}\Bigg).
\end{align*}
Recall that that $\alpha_{t, j}(c_{t,j}) = \min\Big\{\frac{\beta}{n_{t, j}(c_{t,j})}, \half\Big\}$. This implies that if $\alpha_{t, j}(c_{t,j}) = \half$ event $\Epsi_t$ can only hold if $\phat_{t, j}(c_{t,j}) = \half$, in which case we trivially have 
\begin{align*}
    \big(p_{j}(c_{t,j}) - \phat_{t, j}(c_{t,j})\big)\sqrt{\frac{1 - \phat_{t,j}(c_{t,j})}{\phat_{t,j}(c_{t,j})}}  + \big(\phat_{t, j}(c_{t,j}) - p_{j}(c_{t,j})\big)\sqrt{\frac{\phat_{t,j}(c_{t,j})}{1 - \phat_{t,j}(c_{t,j})}} = 0
\end{align*}
We do the remainder of the analysis under the assumptions that $\alpha_{t, j}(c_{t,j}) \not = \half$ and $p_{j}(c_{t,j}) - \phat_{t, j}(c_{t,j}) < 0$. The case where $p_{j}(c_{t,j}) - \phat_{t, j}(c_{t,j}) \geq 0$ follows from symmetrical arguments. We have that 
\begin{align*}
    & \big(p_{j}(c_{t,j}) - \phat_{t, j}(c_{t,j})\big)\sqrt{\frac{1 - \phat_{t,j}(c_{t,j})}{\phat_{t,j}(c_{t,j})}}  + \big(\phat_{t, j}(c_{t,j}) - p_{j}(c_{t,j})\big)\sqrt{\frac{\phat_{t,j}(c_{t,j})}{1 - \phat_{t,j}(c_{t,j})}} \\
    & \leq \big(\phat_{t, j}(c_{t,j}) - p_{j}(c_{t,j})\big)\sqrt{\frac{\phat_{t,j}(c_{t,j})}{1 - \phat_{t,j}(c_{t,j})}} \\
    & \leq \Bigg(\sqrt{\frac{\hat{V}_{t, j}(c_{t,j})}{n_{t, j}(c_{t,j})}2\ln(3/\delta)} + \frac{3\ln(3/\delta)}{n_{t, j}(c_{t,j})}\Bigg)\sqrt{\frac{\phat_{t,j}(c_{t,j})}{1 - \phat_{t,j}(c_{t,j})}},
\end{align*}
where the last inequality holds because of the assumption that $\Lambda_{\delta,t}$ holds. 
Since $\hat{p}_{t, j}(c_{t,j}) \in [\alpha_{t, j}(c_{t,j}), 1-\alpha_{t, j}(c_{t,j})]$ we have that 
\begin{align*}
    &\Bigg(\sqrt{\frac{\phat_{t,j}(c_{t,j})(1-\phat_{t,j}(c_{t,j}))}{n_{t, j}(c_{t,j})}2\ln(3/\delta)} + \frac{3\ln(3/\delta)}{n_{t, j}(c_{t,j})}\Bigg)\sqrt{\frac{\phat_{t,j}(c_{t,j})}{1 - \phat_{t,j}(c_{t,j})}} \\
    & \leq \big(\sqrt{\frac{2}{\beta} \ln(3/\delta)} + \frac{3}{\beta}\ln(3/\delta)\big)\sqrt{{\phat_{t,j}(c_{t,j})}({1 - \phat_{t,j}(c_{t,j})})}
\end{align*}
where we used that $\alpha_{t, j}(c_{t,j}) \leq \frac{\beta}{n_{t, j}(c_{t,j})} \leq 1 - \phat$. The above implies that 
\begin{align*}
    & \E_{Z}[\E_{\yhatt}[\id[\yhatt \not = y_t]]] \leq \half\Big(1 + \half \big(\sqrt{\frac{2}{\beta} \ln(3/\delta)} + \frac{3}{\beta}\ln(3/\delta)\big)\Big)^K\prod_{j = 1}^K 2\sqrt{{\phat_{t,j}(c_{t,j})}({1 - \phat_{t,j}(c_{t,j})})}
\end{align*}
Since $\beta = {18\ln(3/\delta)K^2}$ we have that 
\begin{align*}
    \Big(1 + \half \big(\sqrt{\frac{2}{\beta} \ln(3/\delta)} + \frac{3}{\beta}\ln(3/\delta)\big)\Big)^K \leq  (1 + \frac{1}{3K})^K \leq 1/(1 - 1/3) = \frac{3}{2}
\end{align*}
and thus 
\begin{align*}
    & \E_{Z}[\E_{\yhatt}[\id[\yhatt \not = y_t]]] \leq \frac{3}{4}\prod_{j = 1}^K 2\sqrt{{\phat_{t,j}(c_{t,j})}({1 - \phat_{t,j}(c_{t,j})})} \\
    & = \frac{3}{4} \prod_{j = 1}^K (1 - 4(\half - \phat_{t, j}(c_{t, j}))^2) \leq \frac{3}{4} \exp\Big(-2\sumK(\half - \phat_{t,j}(c_{t,j}))^2\Big)
\end{align*}
where the last inequality is due to $1 + x \leq \exp(x)$. 
\end{proof}

\lemmultiplicativeboundtwo*

\begin{proof}
Denote by $g_t = \exp\Big(-2\sumK(\half - \phat_{t,j}(c_{t,j}))^2\Big)$. Using the fact that $\exp(x)$ and $(\half - x)^2$ are convex in $x$ we can see that
\begin{align*}
    & \frac{3}{4} \exp\Big(-2\sumK(\half - \phat_{t,j}(c_{t,j}))^2\Big) \\
    & = \frac{7}{8}\exp\Big(-2\sumK(\half - p_{j}(c_{t,j}))^2\Big) + \frac{7}{8} \exp\Big(-2\sumK(\half - \phat_{t,j}(c_{t,j}))^2\Big) - \frac{7}{8} \exp\Big(-2\sumK(\half - p_{j}(c_{t,j}))^2\Big) - \frac{1}{8} g_t\\
    & \leq \frac{7}{8} \exp\Big(-2\sumK(\half - p_{j}(c_{t,j}))^2\Big) + \frac{28}{8} g_t \sumK|p_{j}(c_{t,j}) - \phat_{t,j}(c_{t,j})| - \frac{1}{8} g_t \\
    & \leq \frac{7}{8} \exp\Big(-2\sumK(\half - p_{j}(c_{t,j}))^2\Big) + g_t^2/12 + 3 \Big(\frac{28}{8}\Big)^2 K \sumK|p_{j}(c_{t,j}) - \phat_{t,j}(c_{t,j})|^2 - \frac{1}{8} g_t \\
    & \leq \frac{7}{8} \exp\Big(-2\sumK(\half - p_{j}(c_{t,j}))^2\Big) + 3 \Big(\frac{28}{8}\Big)^2 K \sumK|p_{j}(c_{t,j}) - \phat_{t,j}(c_{t,j})|^2,
\end{align*}
where the third inequality is because $ab \leq \frac{a^2}{2\eta} + \frac{\eta}{2}b^2$ for $\eta > 0$ and the last inequality is due to the fact that $g_t \geq g_t^2$. 
Therefore, we have that 
\begin{align}
    & \frac{3}{4} \exp\Big(-2\sumK(\half - \phat_{t,j}(c_{t,j}))^2\Big) \nonumber  \\ 
    & \leq \frac{7}{8} \exp\Big(-2\sumK(\half - p_{j}(c_{t,j}))^2\Big) + 3 \Big(\frac{28}{8}\Big)^2 K \sumK|p_{j}(c_{t,j}) - \phat_{t,j}(c_{t,j})|^2 \nonumber \\
    & \leq \frac{7}{8} \exp\Big(-2\sumK(\half - p_{j}(c_{t,j}))^2\Big) + 6 \Big(\frac{28}{8}\Big)^2 K \sumK \Bigg(\frac{2\ln(3/\delta)}{n_{t,j}(c_{t,j})} + \Big(\frac{3\ln(3/\delta)}{n_{t,j}(c_{t,j})}\Big)^2\Bigg), \nonumber 
\end{align}
where the last inequality is due to the assumption that $\Lambda_{\delta,t}$ holds.

\end{proof}

\lemoptimism* 

\begin{proof}
Since we assume that $\Lambda_{\delta,t}$ holds we have that 
\begin{align*}
    \min_{\c \in \Cshat} \Bigg\{  \exp\Big(-2\sumK(\half - p_{j}(c_{j}))^2\Big) + \lambda \sumK c_{j} \Bigg\} \geq  \exp\Big(-2\sumK(\half - \Phat_{t, j}(c_{t, j}))^2\Big) + \lambda \sumK c_{t, j}.
\end{align*}
Denote by $h_t = \exp\Big(-2\sumK(\half - p_{j}(c_{t, j}))^2\Big)$ By the above inequality and the fact that $\exp(x)$ and $(\half - x)^2$ are convex, we have that 
\begin{align}\label{eq:intermediatebound}
    & \exp\Big(-2\sumK(\half - p_{j}(c_{t,j}))^2\Big) + \lambda \sumK c_{j}  - \min_{\c \in \Cshat}\Bigg\{  \exp\Big(-2\sumK(\half - p_{j}(c_{j}))^2\Big) + \lambda \sumK c_{j} \Bigg\} \nonumber\\
    & \leq \exp\Big(-2\sumK(\half - p_{j}(c_{t,j}))^2\Big) -  \exp\Big(-2\sumK(\half - \Phat_{t, j}(c_{t, j}))^2\Big) \nonumber \\
    & \leq 4 h_t\sumK|\Phat_{t, j}(c_{t, j}) - p_{j}(c_{t,j})| \\
    & \leq \frac{1}{8}h_t^2 + 48 K\sumK|\Phat_{t, j}(c_{t, j}) - p_{j}(c_{t,j})|^2 \nonumber\\
    & \leq \frac{1}{8}h_t^2 + 96K\sumK \Bigg(\frac{2\ln(3/\delta)}{n_{t,j}(c_{t,j})} + \Big(\frac{3\ln(3/\delta)}{n_{t,j}(c_{t,j})}\Big)^2\Bigg) \nonumber
\end{align} 
where the third inequality is because $ab \leq \frac{a^2}{2\eta} + \frac{\eta}{2}b^2$ for $\eta > 0$ and the last inequality is due to the assumption that $\Lambda_{\delta,t}$ holds. Thus, we have that 
\begin{align}\label{eq:boundcostofC}
    & \frac{7}{8}\exp\Big(-2\sumK(\half - p_{j}(c_{t,j}))^2\Big) + \lambda \sumK c_{j} \nonumber \\
    & = \exp\Big(-2\sumK(\half - p_{j}(c_{t,j}))^2\Big) + \lambda \sumK c_{j} - h_t \nonumber \\
    & \leq \min_{\c \in \Cshat}\Bigg\{  \exp\Big(-2\sumK(\half - p_{j}(c_{j}))^2\Big) + \lambda \sumK c_{j} \Bigg\} + 96K\sumK \Bigg(\frac{2\ln(3/\delta)}{n_{t,j}(c_{t,j})} + \Big(\frac{3\ln(3/\delta)}{n_{t,j}(c_{t,j})}\Big)^2\Bigg),
\end{align}
where we used that $h_t^2 \leq h_t$. 

Denote by 
\begin{align*}
    \tilde{\c} & = \argmin_{\c \in \Cshat}\Bigg\{\exp\Big(-2\sumK(\half - p_{j}(c_{j}))^2\Big) + \lambda \sumK c_{j} \Bigg\} \\
    \c^\star & = \argmin_{\c \in [0,1]^K}\Bigg\{\exp\Big(-2\sumK(\half - p_{j}(c_{j}))^2\Big) + \lambda \sumK c_{j} \Bigg\}
\end{align*}
Now, since $\Cshat$ is such that for any $c^\star \in [0,1]$ there is a $\tilde{c} \in \Cshat$ that satisfies $|\tilde{c} - c^\star| \leq \varepsilon$ we have that 
\begin{align}\label{eq:boundcostofCshat}
    & \exp\Big(-2\sumK(\half - p_{j}(\tilde{c}_{j}))^2\Big) + \lambda \sumK \tilde{c}_{j}  - \Big(\exp\Big(-2\sumK(\half - p_{j}(c^\star_{j}))^2\Big) + \lambda \sumK c^\star_{j}\Big) \nonumber\\
    & \leq 4\sumK|p_{j}(\tilde{c}_{j}) - p_{j}(c^\star_{j})|+ \lambda \sumK |\tilde{c}_{j} - c^\star_{j}| \nonumber \\
    & \leq (4L + \lambda) \sumK |\tilde{c}_{j} - c^\star_{j}| \leq (4L + \lambda)K\varepsilon,
\end{align}
where in the first inequality we used the same bound as we used in equation~\eqref{eq:intermediatebound}.
By combining equations \eqref{eq:boundcostofC} and \eqref{eq:boundcostofCshat} we can see that 
\begin{align*}
     \frac{7}{8}\exp\Big(-2\sumK(\half - p_{j}(c_{t,j}))^2\Big) & + \lambda \sumK c_{j} \leq \min_{\c \in [0,1]^K}\Bigg\{\exp\Big(-2\sumK(\half - p_{j}(c_{j}))^2\Big) + \lambda \sumK c_{j} \Bigg\} \\ 
    & \quad + (4L + \lambda)K\varepsilon + 96 K\sumK \Bigg(\frac{2\ln(3/\delta)}{n_{t,j}(c_{t,j})} 
     + \Big(\frac{3\ln(3/\delta)}{n_{t,j}(c_{t,j})}\Big)^2\Bigg)
\end{align*}
\end{proof}

\lemnotinrange* 

\begin{proof}
We work in the case where $\hat{p}_{t, j_t}(c_{t,j_t}) \leq \alpha_{t, j_t}(c_{t,j_t})$. The case where $1 - \hat{p}_{t, j_t}(c_{t,j_t}) \leq \alpha_{t, j_t}(c_{t,j_t})$ follows from symmetric arguments. Since we still work in the case where $\Lambda_{\delta,t}$ holds we have that
\begin{align*}
    & \E_{Z}[\id[\yhatt \not = y_t]] + \lambda \sumK c_{t, j}  = p_{j_t}(c_{t, j_t}) + \lambda \sumK c_{t, j}\nonumber\\
    & \leq \hat{p}_{t, j_t}(c_{t,j_t}) + |p_{j_t}(c_{t, j_t})-\hat{p}_{t, j_t}(c_{t,j_t})| \nonumber + \lambda \sumK c_{t, j}\\
    & \leq \alpha_{t, j_t}(c_{t,j_t}) + \sqrt{\frac{\hat{p}_{t, j_t}(c_{t,j_t})}{n_{t, j}(c_{t,j_t})}2\ln(3/\delta)} + \frac{3\ln(3/\delta)}{n_{t, j_t}(c_{t,j_t})} \nonumber + \lambda \sumK c_{t, j}\\
    & \leq \alpha_{t, j_t}(c_{t,j_t}) + \sqrt{\frac{\alpha_{t, j_t}(c_{t,j_t})}{n_{t, j}(c_{t,j_t})}2\ln(3/\delta)} + \frac{3\ln(3/\delta)}{n_{t, j_t}(c_{t,j_t})} \nonumber + \lambda \sumK c_{t, j}\\
    & = \frac{\beta + 3\ln(3/\delta) + \sqrt{2 \beta \ln(3/\delta)}}{n_{t, j}(c_{t,j_t})} \nonumber + \lambda \sumK c_{t, j} \\
    & \leq \frac{2\beta + 4\ln(3/\delta)}{n_{t, j_t}(c_{t,j_t})} + \lambda \sumK c_{t, j} \nonumber \\
    & \leq \min_{\c \in [0,1]^K}\Bigg\{\exp\Big(-2\sumK(\half - p_{j}(c_{j}))^2\Big) + \lambda \sumK c_{j} \Bigg\} \nonumber \\ 
    & \quad + \frac{2\beta + 4\ln(3/\delta)}{n_{t, j_t}(c_{t,j_t})} + (4L + \lambda)K\varepsilon + 96 K\sumK \Bigg(\frac{2\ln(3/\delta)}{n_{t,j}(c_{t,j})} 
     + \Big(\frac{3\ln(3/\delta)}{n_{t,j}(c_{t,j})}\Big)^2\Bigg),
\end{align*}
where the second inequality follows from the assumption that $\Lambda_{\delta,t}$ holds, the third inequality follows from $\hat{p}_{t, j_t}(c_{t,j_t}) \leq \alpha_{t, j_t}(c_{t,j_t})$, and the final inequality follows from adding $\frac{7}{8}\Phi(\p(\c_t))$  and Lemma~\ref{lem:multiplicativebound2}. 
\end{proof}

\begin{theorem} 
\label{th:maindetail}
    Let $\Cshat$ be such that for any $c^\star \in [0,1]$ there is a $\tilde{c} \in \Cshat$ that satisfies $|\tilde{c} - c^\star| \leq \varepsilon$ and let $\beta = {18\ln(3/\delta)K^2}$. Then for any $\delta \in (0, 1)$
    \begin{align*}
    & \sumT \E_{Z}\Big[\E_{\yhatt}[\id[\yhatt \not = y_t]] + \lambda \sumK c_{t, j}\Big] \\
    & \leq T \min_{\c \in [0,1]^K}\Bigg\{\exp\Big(-2\sumK(\half - p_{j}(c_{j}))^2\Big) + \lambda \sumK c_{j} \Bigg\} + (1 + \lambda K)T^3K\delta + T (4L + \lambda)K\varepsilon\\
    & \quad + N\Big(1 + \lambda K + 2592 K^2\big(\ln(3/\delta)\big)^2 + (1 + \log(T))\Big(K(2\beta + 4\ln(3/\delta)) + 576 K^2 \ln(3/\delta)\big) \Big)\Big) 
    \end{align*}
\end{theorem}

\begin{proof}

First, since $\id[\yhatt \neq y_t] + \lambda \sumK c_{t,j} \leq 1 + K \lambda$ we have that 
\begin{align*}
    \E\left[\sumT \id[\yhatt \not = y_t] + \sumK c_{t,j}\right] \leq N(1 + K \lambda) + \E\left[\sumTN \id[\yhatt \not = y_t]+ \sumK c_{t,j}\right].
\end{align*}
By the Tower rule we have that %
\begin{align*}
    & \E\left[\sumTN \id[\yhatt \not = y_t] + \sumK c_{t,j}\right] = \E\left[\sumTN \E \left[\id[\yhatt \not = y_t] + \sumK c_{t,j}\middle|\Fset_t \right] \right]\\
    & = \E\left[\sumTN \id[\Lambda_{\delta, t}] \E\left[\id[\yhatt \not = y_t] + \sumK c_{t,j}\middle|\Fset_t \right] \right]
    + \E\left[\sumTN \id[\Lambda_{\delta, t}^c] \E\left[\id[\yhatt \not = y_t] + \sumK c_{t,j}\middle|\Fset_t \right] \right] \\
    & \leq \E\left[\sumTN \id[\Lambda_{\delta, t}] \E\left[\id[\yhatt \not = y_t] + \sumK c_{t,j}\middle|\Fset_t\right]\right] + \delta K N T^2 (\lambda K+1),
\end{align*}
where the last inequality follows from Lemma~\ref{lem:difcontrol}. 
The remainder of the proof consists of controlling the conditional expectation on the r.h.s. of the above equation. In the remainder of the proof we assume $\Lambda_{\delta, t}$. By Lemmas \ref{lem:multiplicativebound1}, \ref{lem:multiplicativebound2}, \ref{lem:optimism}, and \ref{lem:notinrange} we have that 
\begin{align*}
    \E[\id[\yhatt \not = y_t] + \sumK c_{t,j}\mid \Fset_t] 
    & \leq \min_{\c \in [0,1]^K}\Bigg\{\Phi(\p(\c)) + \lambda \sumK c_{j} \Bigg\} \nonumber \\ 
    & \quad + \frac{2\beta + 4\ln(3/\delta)}{n_{t, j_t}(c_{t,j_t})} + (4L + \lambda)K\varepsilon + 288 K\sumK \Bigg(\frac{2\ln(3/\delta)}{n_{t,j}(c_{t,j})} 
    + \Big(\frac{3\ln(3/\delta)}{n_{t,j}(c_{t,j})}\Big)^2\Bigg)
\end{align*}
For $t \geq N = |\Cshat|$ we have that $n_{t, j}(c) \geq 1$. We continue by bounding 
\begin{align*}
    & \sum_{t=N+1}^T \sumK \Bigg(\frac{2\ln(3/\delta)}{n_{t,j}(c_{t,j})} 
    + \Big(\frac{3\ln(3/\delta)}{n_{t,j}(c_{t,j})}\Big)^2\Bigg) \\
    & = \sum_{t=N+1}^T \sumK \sum_{c \in \Cshat} \id[c_{t,j} = c]\Bigg(\frac{2\ln(3/\delta)}{n_{t,j}(c)} 
    + \Big(\frac{3\ln(3/\delta)}{n_{t,j}(c)}\Big)^2\Bigg) \\
    & \leq K N \big((1 + \log(T))2\ln(3/\delta) + 9\big(\ln(3/\delta)\big)^2\big) ,
\end{align*}
where the inequality is $\sum_{t=1}^T \frac{1}{t} \leq 1 + \log(T)$ and $\sum_{t=1}^T \frac{1}{t^2} \leq 2$.
Similarly, we have 
\begin{align*}
    \sum_{t=N+1}^T \frac{2\beta + 4\ln(3/\delta)}{n_{t, j_t}(c_{t,j_t})} \leq & K N (2\beta + 4\ln(3/\delta))(1 + \log(T))
\end{align*}
Thus, combining the above equations we obtain 
\begin{align*}
    & \E\left[\sumT \id[\yhatt \not = y_t]] + \lambda \sumK c_{t, j}\right] \\
    & \leq T \min_{\c \in [0,1]^K}\Bigg\{\exp\Big(-2\sumK(\half - p_{j}(c_{j}))^2\Big) + \lambda \sumK c_{j} \Bigg\} + N(1 + \lambda K) + (1 + \lambda K)NT^2K\delta\nonumber \\ 
    & \quad + K N (2\beta + 4\ln(3/\delta))(1 + \log(T)) + T (4L + \lambda)K\varepsilon + 288 K^2 N \big((1 + \log(T))2\ln(3/\delta) + 9\big(\ln(3/\delta)\big)^2\big)\\
    & = T \min_{\c \in [0,1]^K}\Bigg\{\exp\Big(-2\sumK(\half - p_{j}(c_{j}))^2\Big) + \lambda \sumK c_{j} \Bigg\} + (1 + \lambda K)NT^2K\delta + T (4L + \lambda)K\varepsilon\\
    & \quad + N\Big(1 + \lambda K + 2592 K^2\big(\ln(3/\delta)\big)^2 + (1 + \log(T))\Big(K(2\beta + 4\ln(3/\delta)) + 576 K^2 \ln(3/\delta)\big) \Big)\Big),
\end{align*}
which completes the proof after replacing $\beta = 8\ln(3/\delta)K^2$.
\end{proof}
\newpage

\section{ADDITIONAL EXPERIMENTS}\label{app:experiments}
Here we present the additional experimental results announced in Section~\ref{sec:experiments}. 

\begin{figure}[H]
    \centering
        \begin{subfigure}
         \centering
         \includegraphics[width=0.45\textwidth]{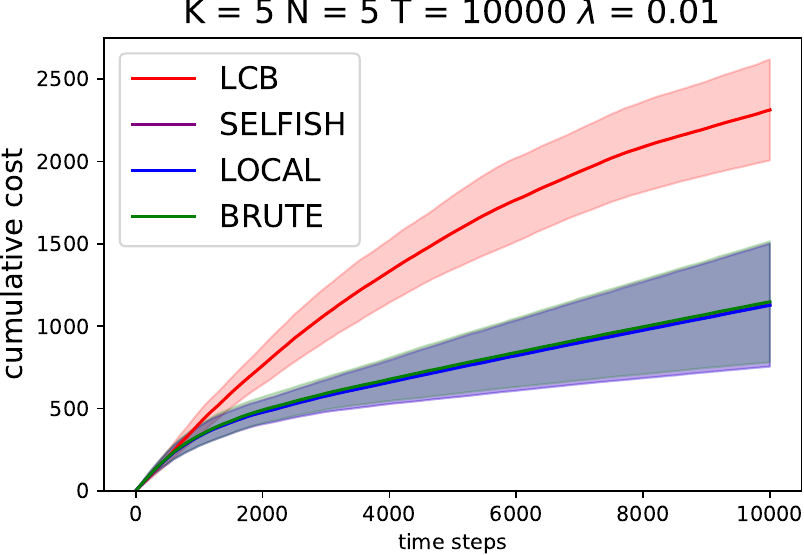}
     \end{subfigure}
    \begin{subfigure}
         \centering
         \includegraphics[width=0.45\textwidth]{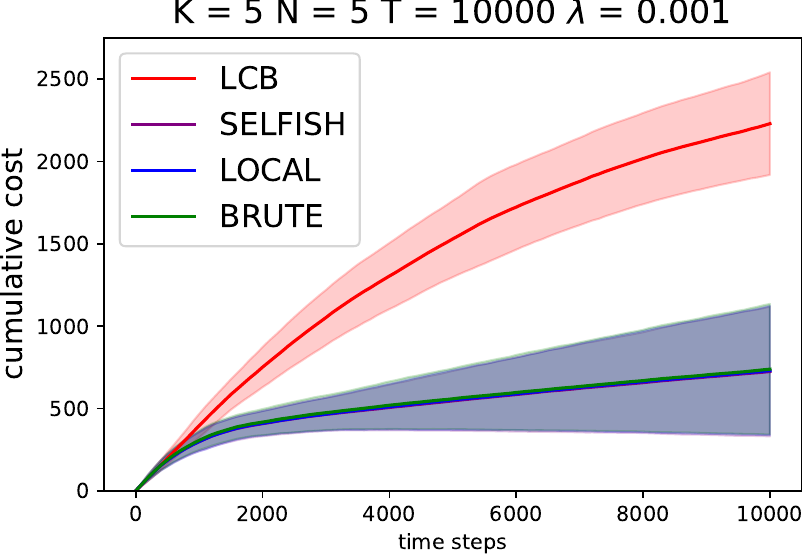}
     \end{subfigure}
    \begin{subfigure}
         \centering
         \includegraphics[width=0.45\textwidth]{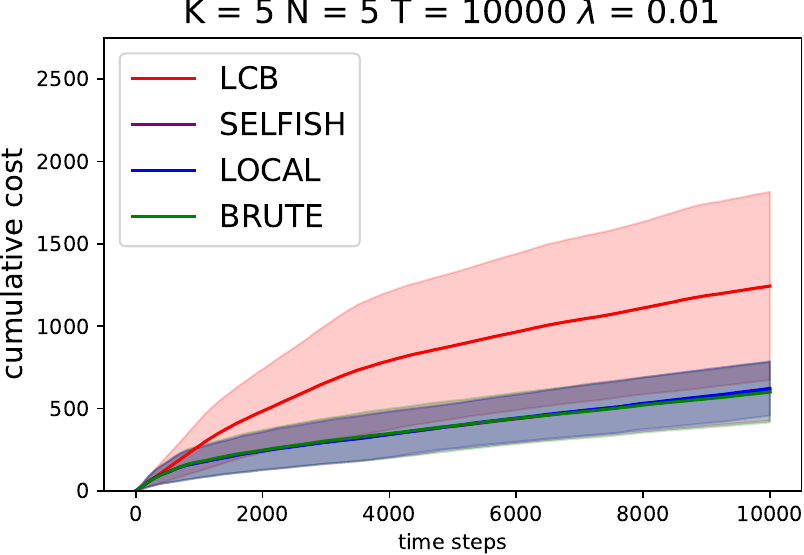}
     \end{subfigure}
     \begin{subfigure}
         \centering
         \includegraphics[width=0.45\textwidth]{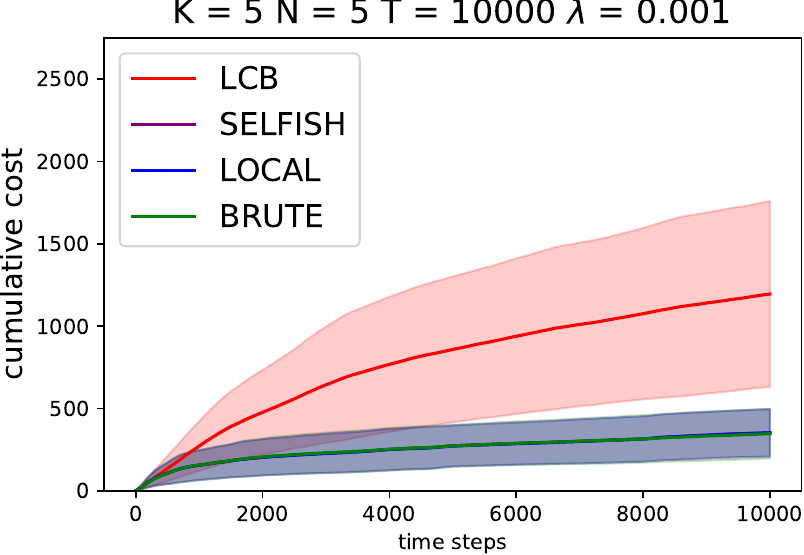}
     \end{subfigure}

    \caption{The top two figures and the bottom two figures show the performance of algorithms on the first and second productivity function, respectively.}
    
    \label{fig: brute}
\end{figure}

\begin{figure}[H]
    \centering
        \begin{subfigure}
         \centering
         \includegraphics[width=0.45\textwidth]{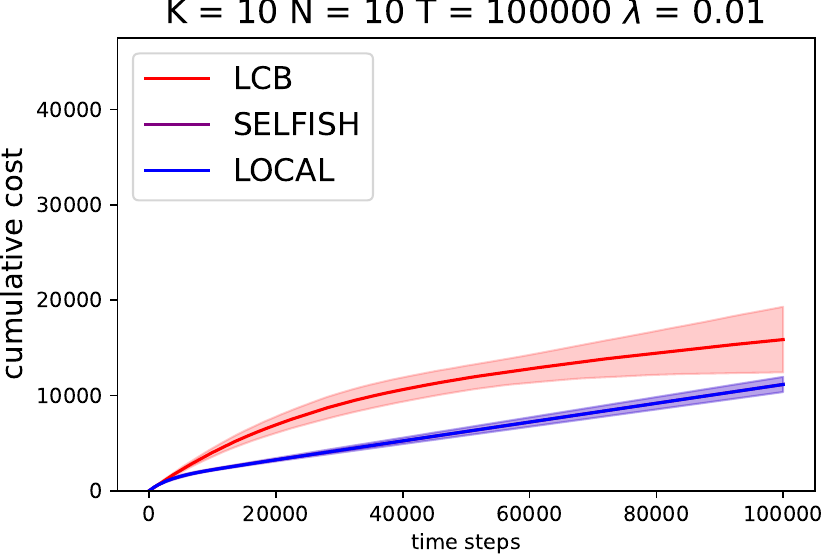}
     \end{subfigure}
    \begin{subfigure}
         \centering
         \includegraphics[width=0.45\textwidth]{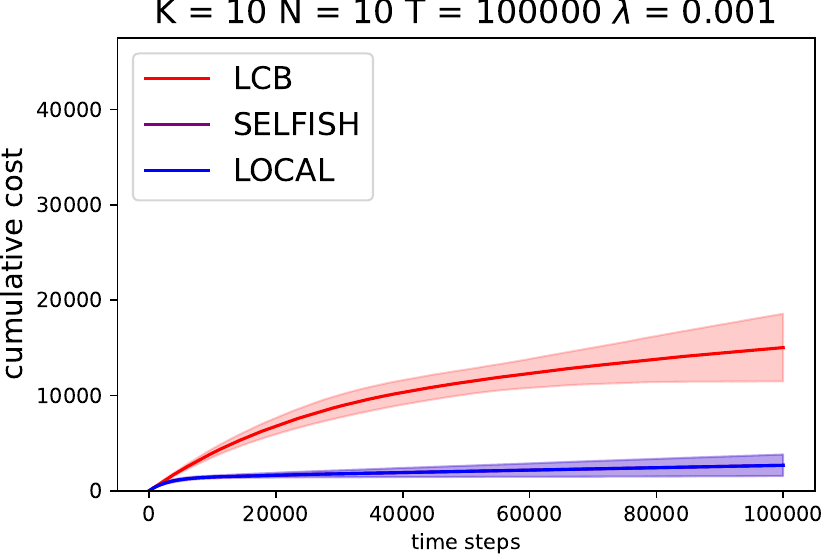}
     \end{subfigure}
    \begin{subfigure}
         \centering
         \includegraphics[width=0.45\textwidth]{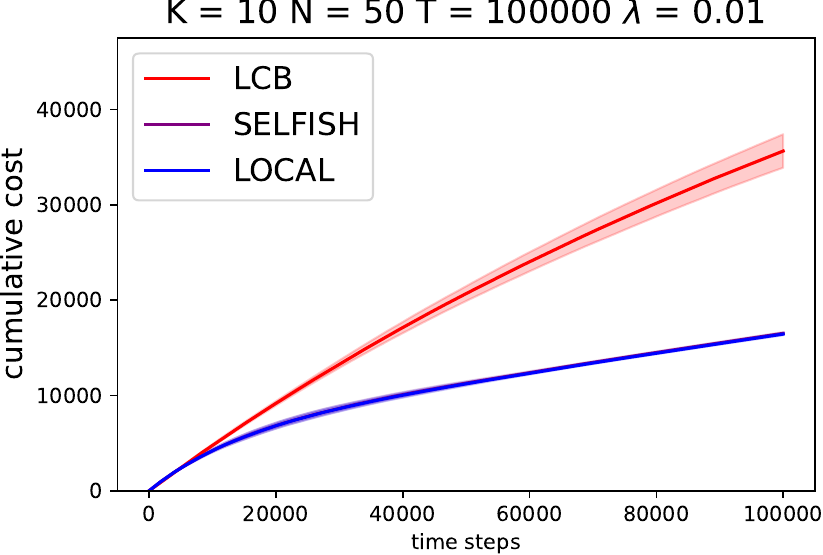}
     \end{subfigure}
     \begin{subfigure}
         \centering
         \includegraphics[width=0.45\textwidth]{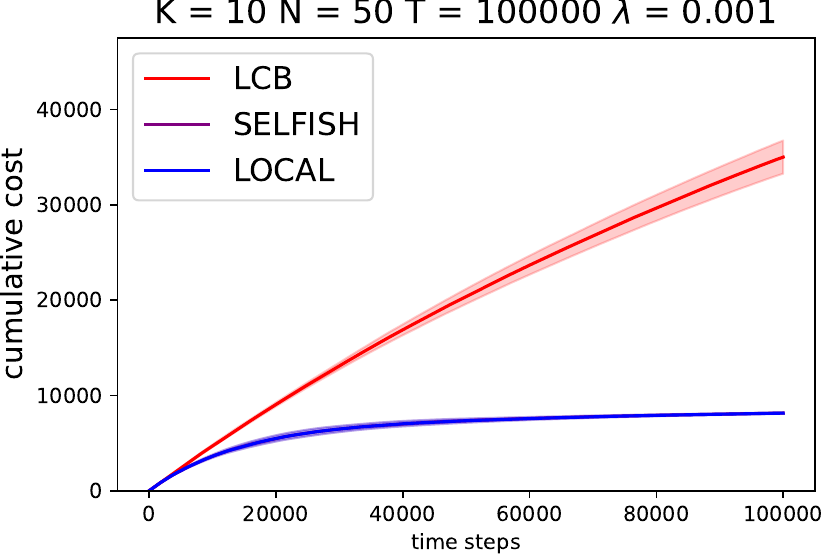}
     \end{subfigure}
    \begin{subfigure}
         \centering
         \includegraphics[width=0.45\textwidth]{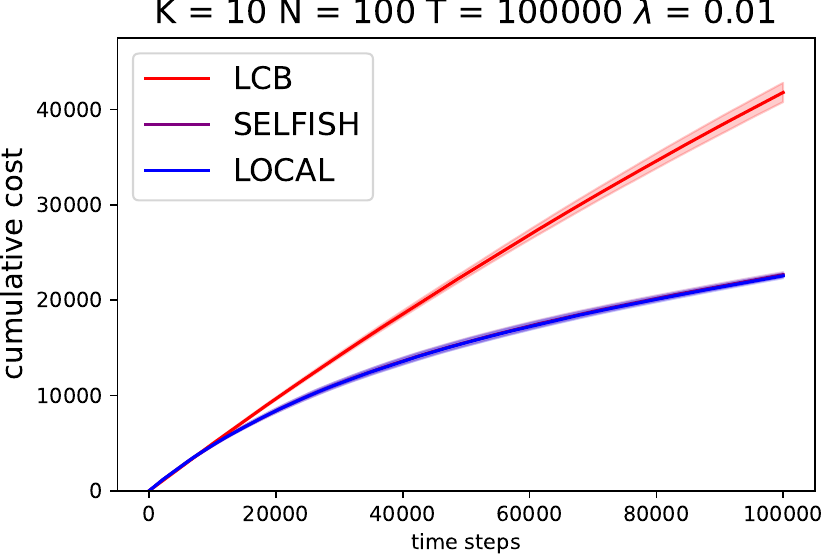}
     \end{subfigure}
    \begin{subfigure}
         \centering
         \includegraphics[width=0.45\textwidth]{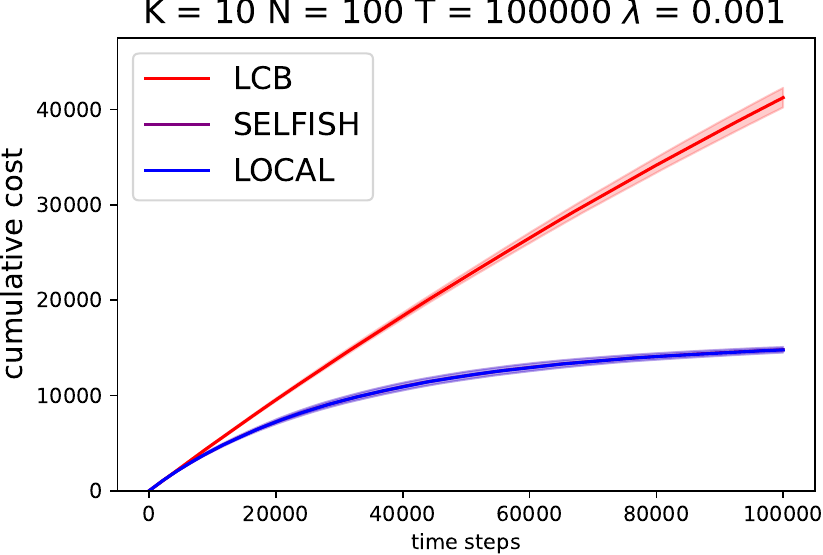}
     \end{subfigure}

    \caption{Plots showing cumulative cost when the algorithms are using the first productivity function for $T = 10^5$, $K=10$, $N \in \{10,50,100\}$, and $\lambda \in \big\{10^{-2},10^{-3}\big\}$.}
    \label{fig:nobrute_K10_func1}
\end{figure}

\begin{figure}[H]
    \centering
        \begin{subfigure}
         \centering
         \includegraphics[width=0.45\textwidth]{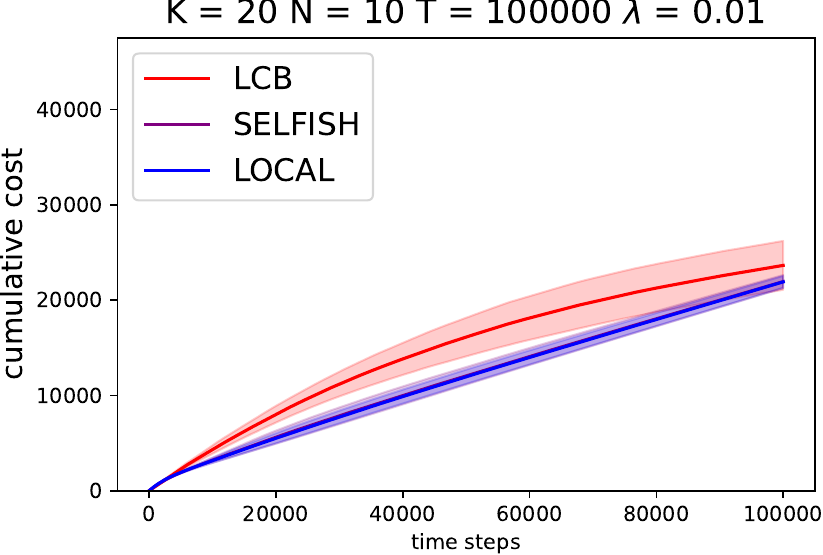}
     \end{subfigure}
    \begin{subfigure}
         \centering
         \includegraphics[width=0.45\textwidth]{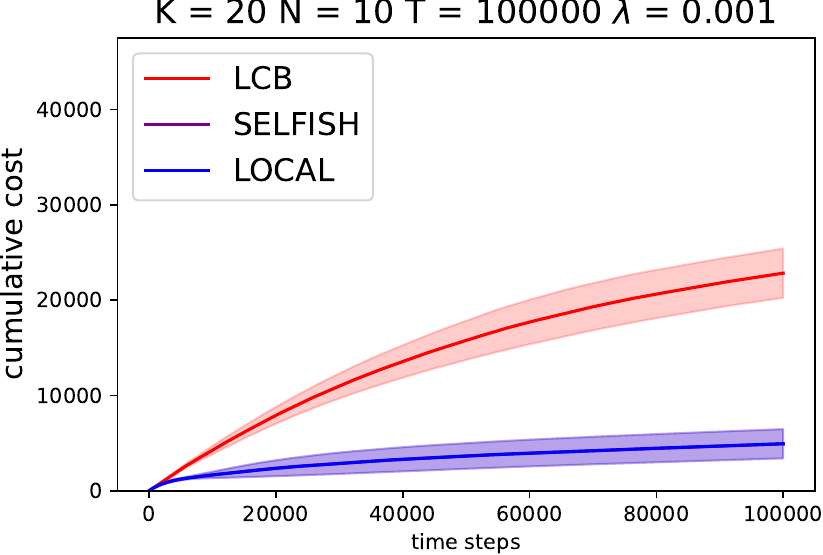}
     \end{subfigure}
    \begin{subfigure}
         \centering
         \includegraphics[width=0.45\textwidth]{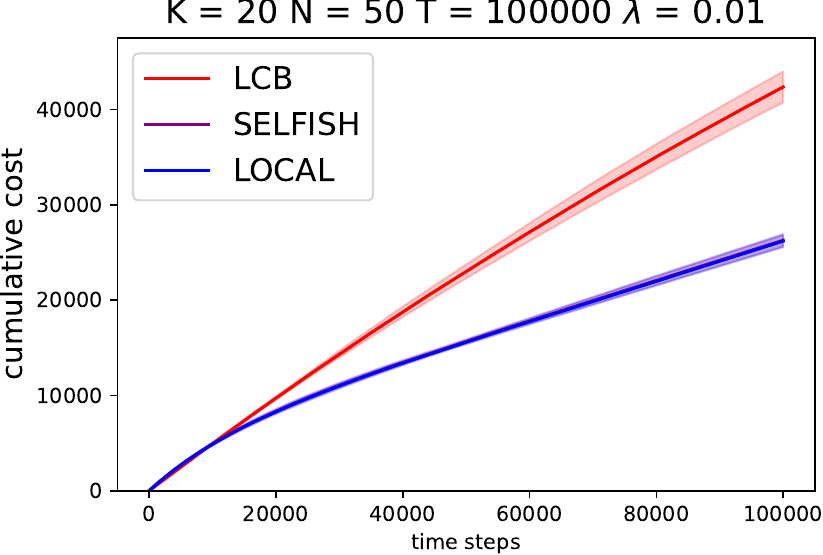}
     \end{subfigure}
     \begin{subfigure}
         \centering
         \includegraphics[width=0.45\textwidth]{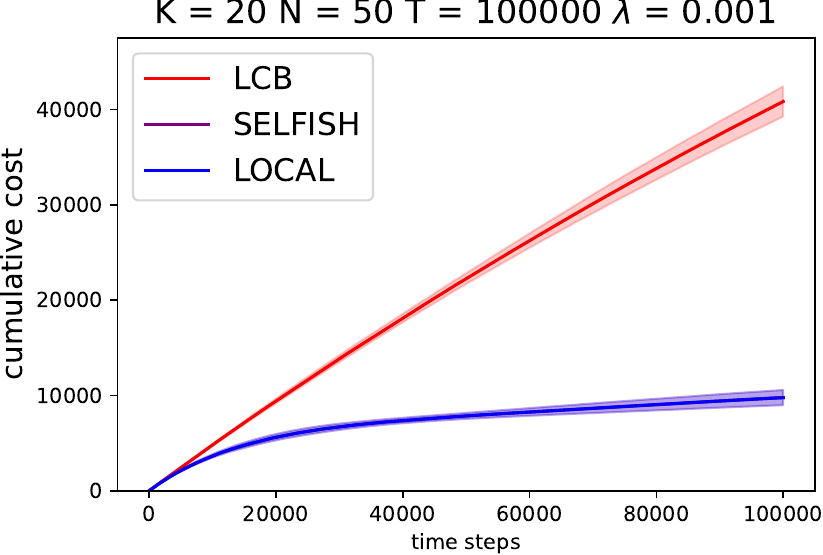}
     \end{subfigure}
    \begin{subfigure}
         \centering
         \includegraphics[width=0.45\textwidth]{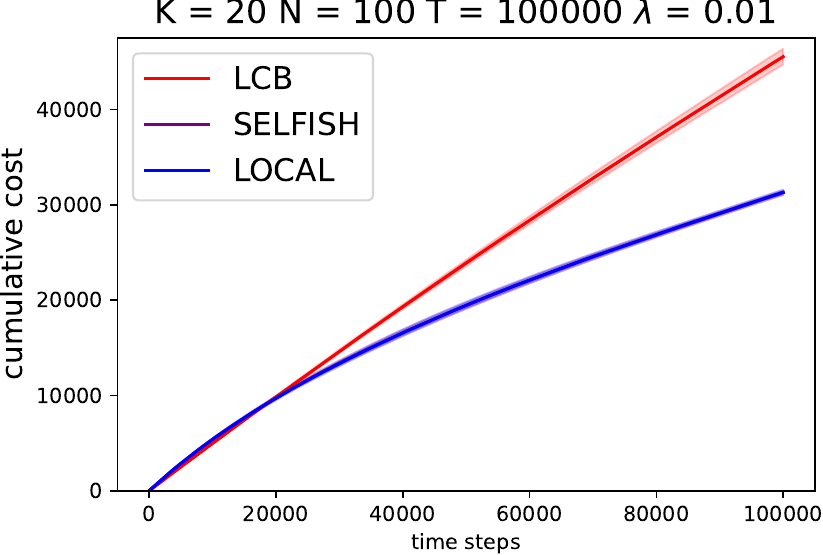}
     \end{subfigure}
    \begin{subfigure}
         \centering
         \includegraphics[width=0.45\textwidth]{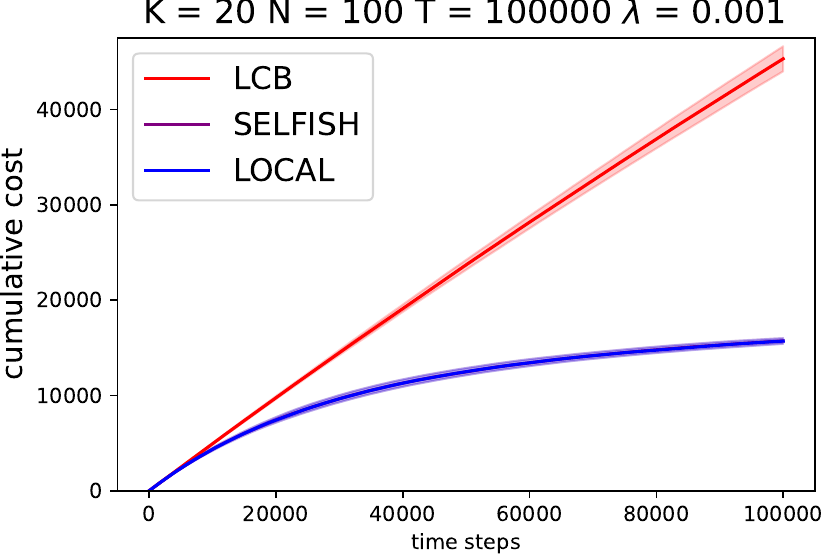}
     \end{subfigure}
    \caption{Plots showing cumulative cost when the algorithms are using the first productivity function for $T = 10^5$, $K=20$, $N \in \{10,50,100\}$, and $\lambda \in \big\{10^{-2},10^{-3}\big\}$.}
    \label{fig:nobrute_K20_func1}
\end{figure}

\begin{figure}[H]
        \centering
        \begin{subfigure}
         \centering
         \includegraphics[width=0.45\textwidth]{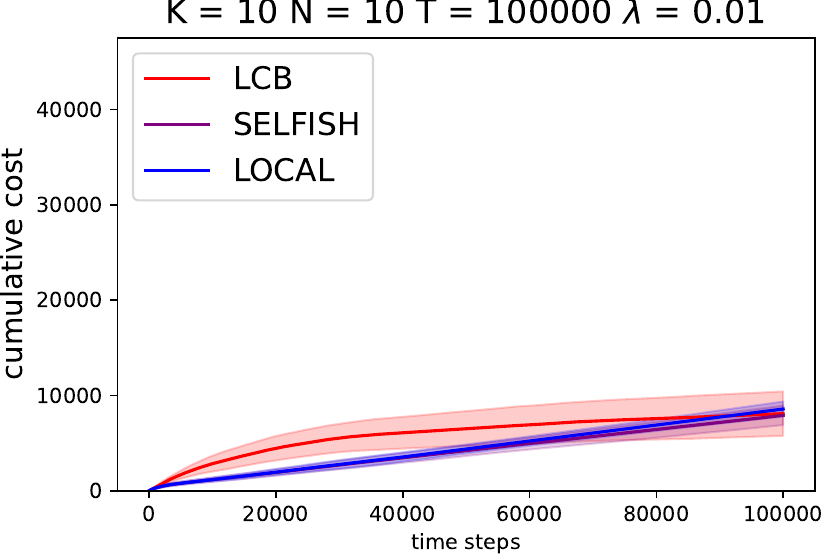}
     \end{subfigure}
    \begin{subfigure}
         \centering
         \includegraphics[width=0.45\textwidth]{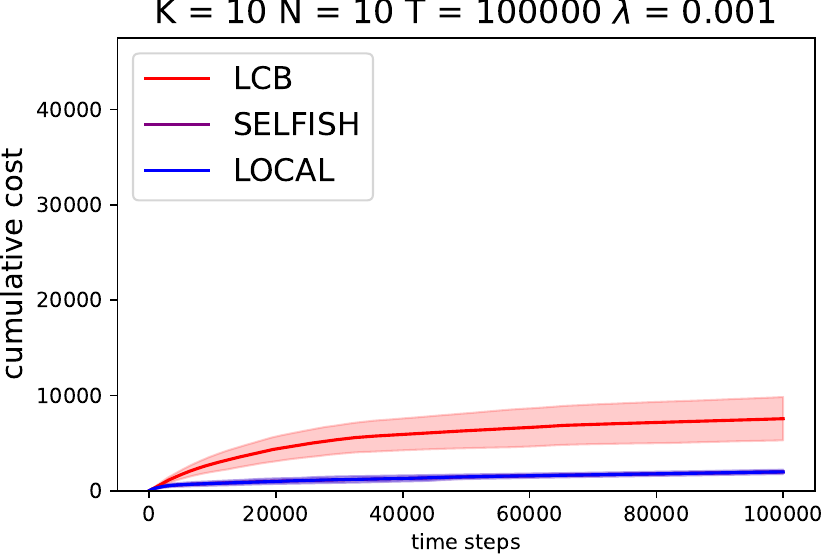}
     \end{subfigure}
    \begin{subfigure}
         \centering
         \includegraphics[width=0.45\textwidth]{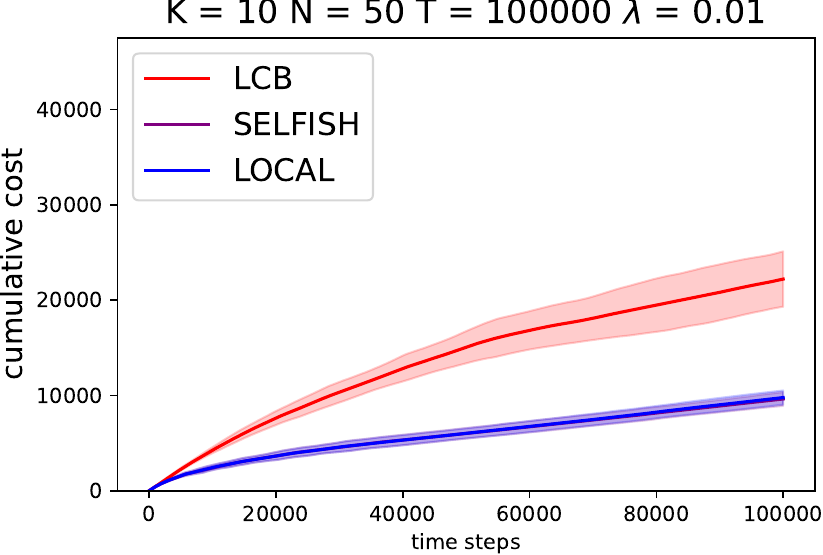}
     \end{subfigure}
     \begin{subfigure}
         \centering
         \includegraphics[width=0.45\textwidth]{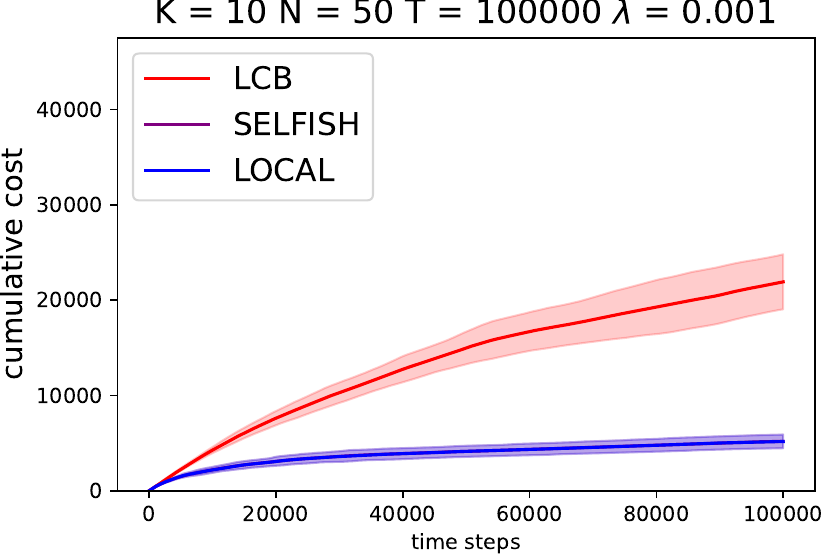}
     \end{subfigure}
    \begin{subfigure}
         \centering
         \includegraphics[width=0.45\textwidth]{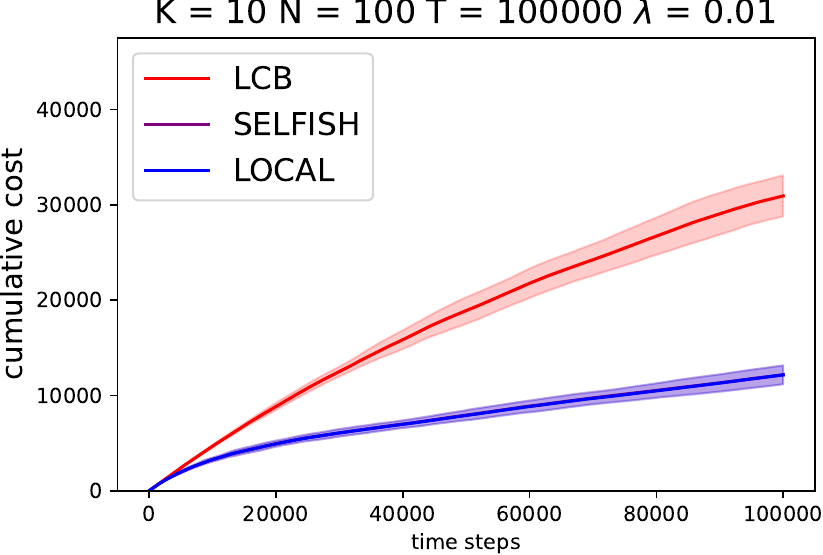}
     \end{subfigure}
    \begin{subfigure}
         \centering
         \includegraphics[width=0.45\textwidth]{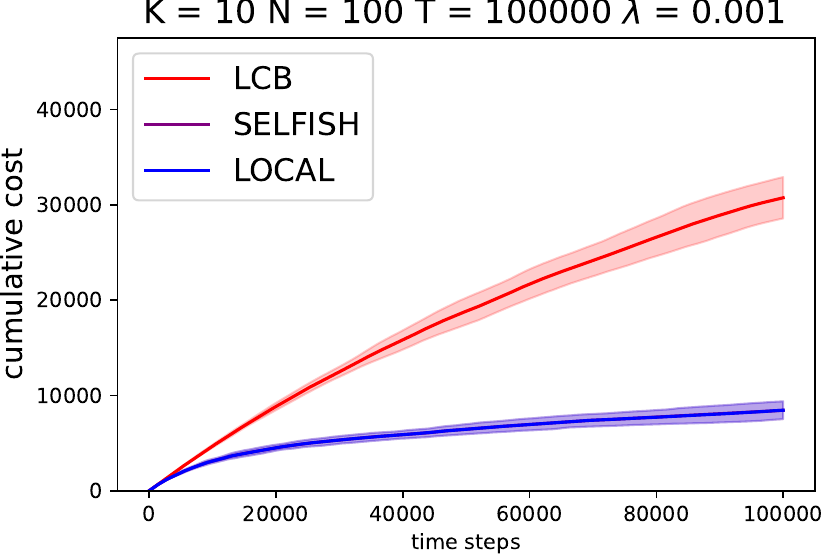}
     \end{subfigure}
     
    \caption{Plots showing cumulative cost when the algorithms are using the second productivity function for $T = 10^5$, $K=10$, $N \in \{10,50,100\}$, and $\lambda \in \big\{10^{-2},10^{-3}\big\}$.}
    \label{fig:nobrute_K10_func2}
\end{figure}

\begin{figure}[H]
    \centering
        \begin{subfigure}
         \centering
         \includegraphics[width=0.45\textwidth]{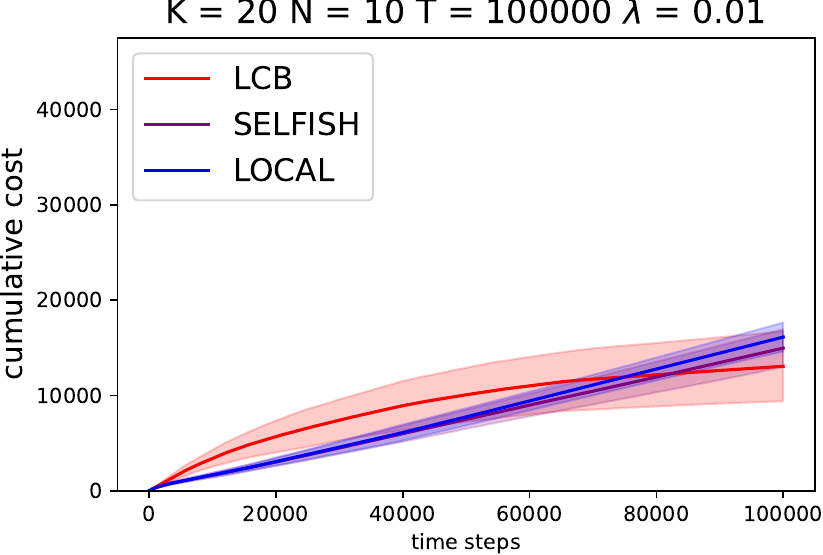}
     \end{subfigure}
    \begin{subfigure}
         \centering
         \includegraphics[width=0.45\textwidth]{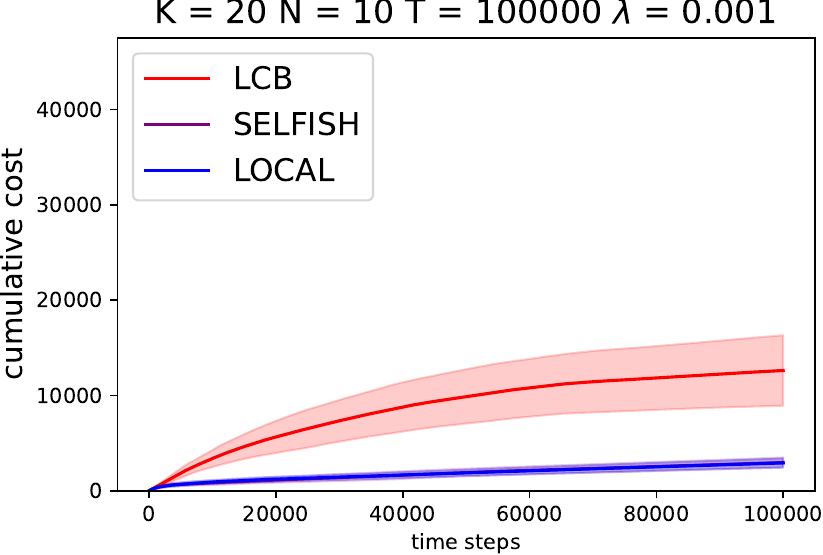}
     \end{subfigure}
    \begin{subfigure}
         \centering
         \includegraphics[width=0.45\textwidth]{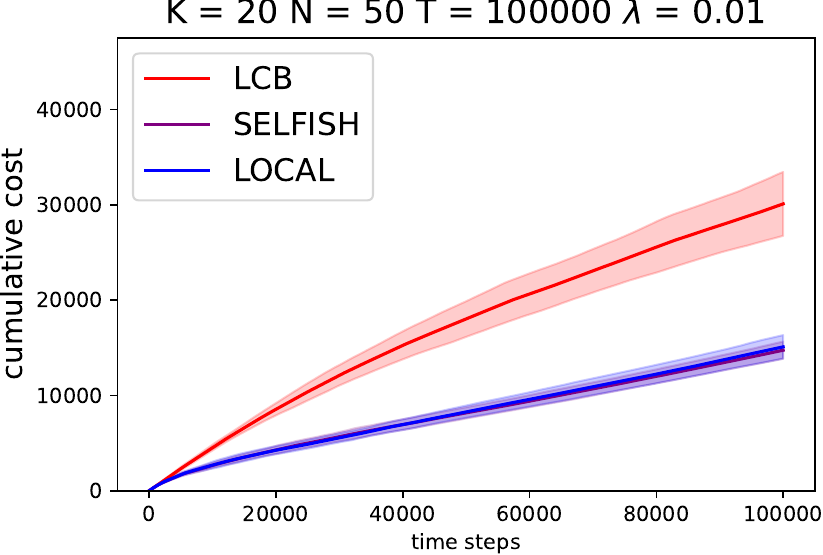}
     \end{subfigure}
     \begin{subfigure}
         \centering
         \includegraphics[width=0.45\textwidth]{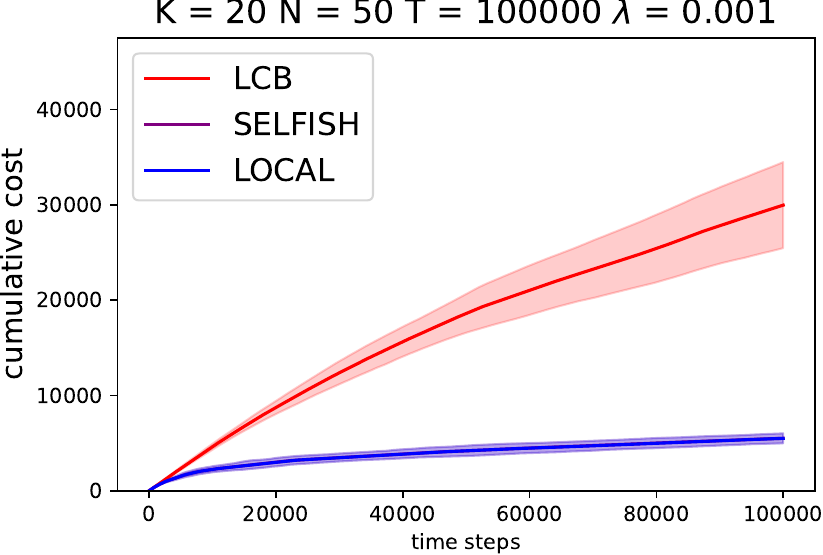}
     \end{subfigure}
    \begin{subfigure}
         \centering
         \includegraphics[width=0.45\textwidth]{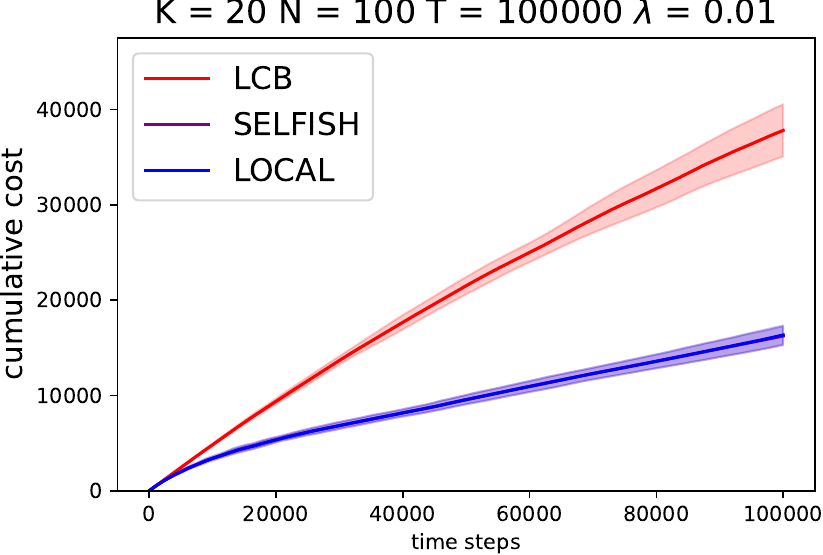}
     \end{subfigure}
    \begin{subfigure}
         \centering
         \includegraphics[width=0.45\textwidth]{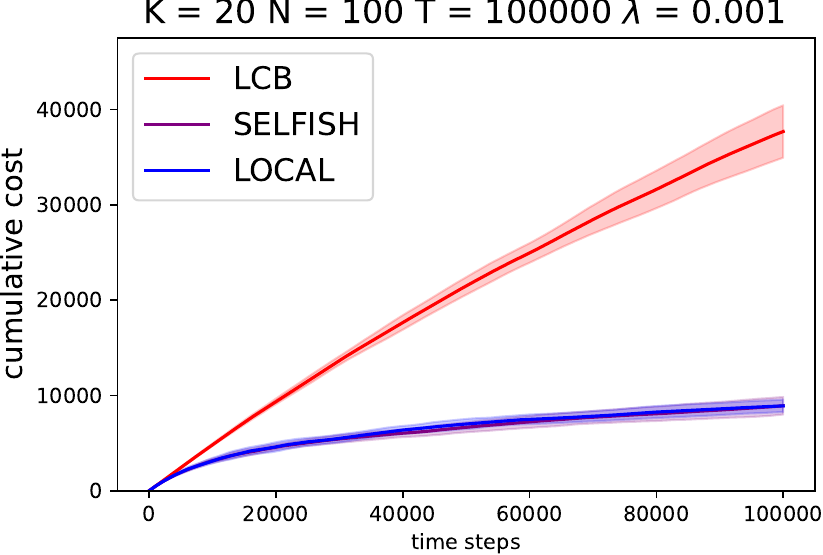}
     \end{subfigure}

    \caption{Plots showing cumulative cost when the algorithms are using the second productivity function for $T = 10^5$, $K=20$, $N \in \{10,50,100\}$, and $\lambda \in \big\{10^{-2},10^{-3}\big\}$.}
    \label{fig:nobrute_K20_func2}
\end{figure}

\end{document}